\documentclass[10pt,twocolumn,letterpaper]{article}

\usepackage[pagenumbers]{wacv} 

\usepackage[normalem]{ulem} 

\usepackage{multirow}
\usepackage{tikz}
\usetikzlibrary{arrows.meta, positioning, calc, fit, backgrounds, decorations.pathreplacing, shadows}
\usepackage{pgfplots}
\pgfplotsset{compat=1.18}

\definecolor{wacvblue}{rgb}{0.21,0.49,0.74}
\usepackage[pagebackref,breaklinks,colorlinks,allcolors=wacvblue]{hyperref}

\usepackage[capitalize,noabbrev]{cleveref}

\def\wacvPaperID{1573} 
\def\confName{WACV}
\def\confYear{2027}

\title{LipSSD: Lipschitz-Constrained Single-Shot Detection for Adversarially Robust Object Detection}

\author{
Vincent Lébé$^{1,2}$ \quad Yannick Prudent$^{1}$ \quad Corentin Friedrich$^{1}$\\[-1pt]
Thomas Massena$^{3,4}$ \quad Ronan Sicre$^{4}$ \quad Franck Mamalet$^{1}$\\[-1pt]
{\small $^{1}$IRT Saint-Exupéry \quad $^{2}$Alstom \quad $^{3}$SNCF, DTIPG \quad $^{4}$IRIT}\\[-1pt]
{\scriptsize\ttfamily
\resizebox{0.94\textwidth}{!}{%
vincent.lebe@alstomgroup.com \quad thomas.massena@sncf.fr \quad ronan.sicre@irit.fr \quad
\{yannick.prudent,corentin.friedrich,franck.mamalet\}@irt-saintexupery.com%
}}
}

\begin{document}
\maketitle
\begin{abstract}
    Object detectors have many applications in safety-critical systems, but they are
    known to be sensitive to worst-case perturbations such as adversarial attacks, which
    limits their applicability in real-world scenarios. Compared with classification,
    adversarial robustness for object detection has received less attention, and
    existing methods are often tied to adversarial training, whose performance may not
    transfer across attacks, perturbation budgets, or architectures. In this work, we
    introduce Lipschitz-constrained variants of object detection architectures as
    robust-by-design alternatives to standard detectors. We validate this approach with
    LipSSD, a Lipschitz-constrained Single Shot MultiBox Detector (SSD), and provide a
    comprehensive study of its adversarial robustness using multiple white-box
    adversarial attacks and datasets. We first analyze the accuracy-robustness trade-off
    induced by Lipschitz constraints and show that it can be controlled through a single
    training hyperparameter. We then demonstrate that Lipschitz-constrained detectors
    are complementary to adversarial training: under the same training setup on the Pascal VOC dataset,
    adversarially trained LipSSD improves mAP@50 on unseen attacks by up to $15$ points
    over classical adversarially trained SSD. Finally, we use more specific
    safety-critical datasets such as LARD and KITTI, and show that Lipschitz-constrained
    detectors can improve robustness while largely preserving clean performance. These
    results suggest that architectural Lipschitz control is a practical and
    attack-agnostic direction for improving the robustness of object detectors.
\end{abstract}
    
\section{Introduction}
\label{sec:intro}

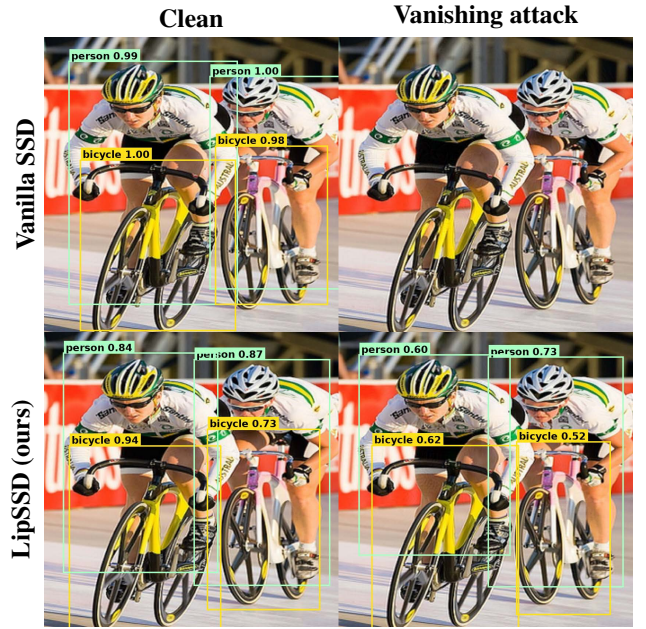
\begin{figure}[t]
    \centering
    \resizebox{\linewidth}{!}{%
    \begin{tikzpicture}[font=\sffamily]
        \node[inner sep=0] (img)
            {\includegraphics[width=8cm]{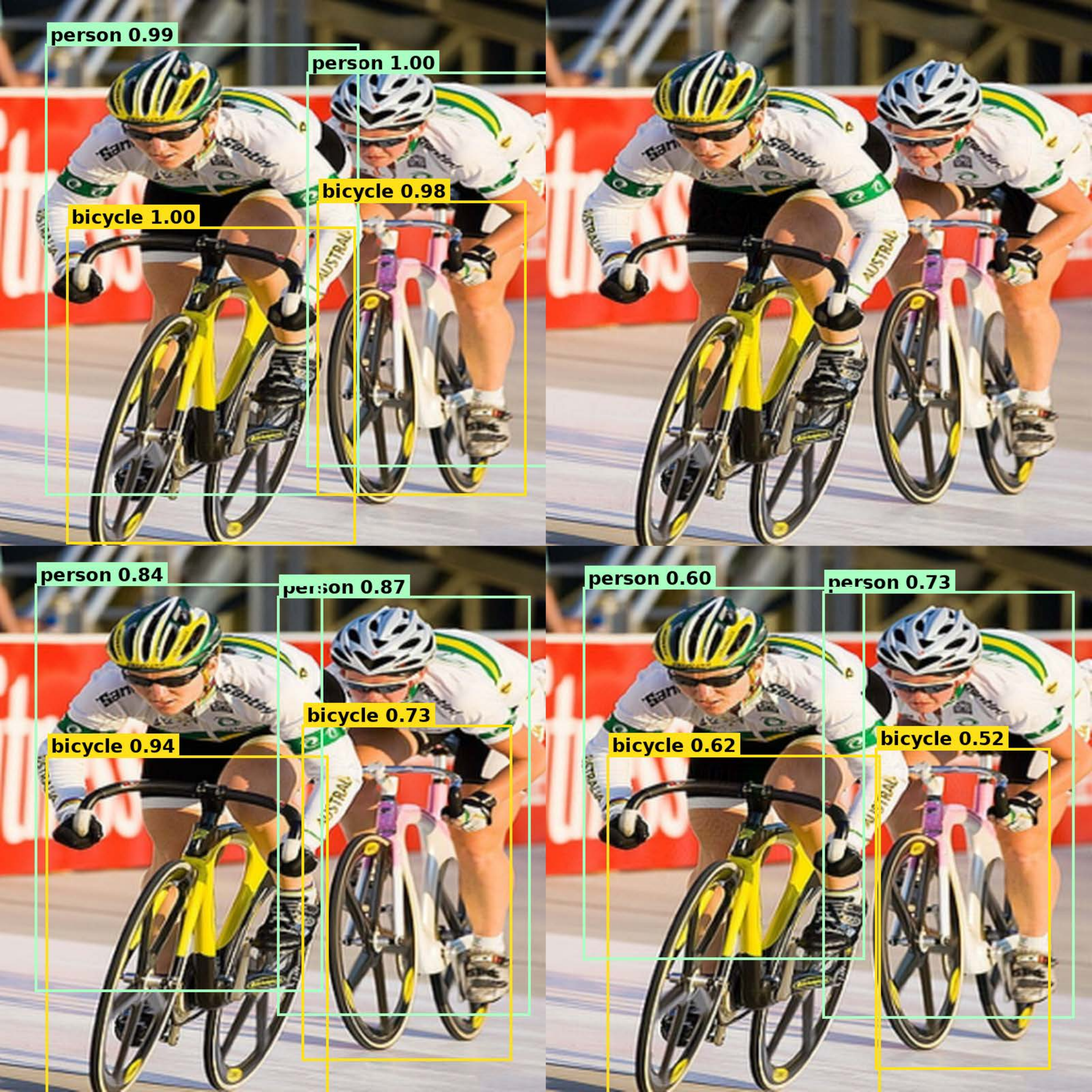}};
        \node[anchor=south, font=\bfseries]
            at ($(img.north west)!0.25!(img.north east)$) {Clean};
        \node[anchor=south, font=\bfseries]
            at ($(img.north west)!0.75!(img.north east)$) {Vanishing attack};
        \node[anchor=south, rotate=90, font=\bfseries]
            at ($(img.north west)!0.25!(img.south west)$) {Vanilla SSD};
        \node[anchor=south, rotate=90, font=\bfseries]
            at ($(img.north west)!0.75!(img.south west)$) {LipSSD (ours)};
    \end{tikzpicture}}
    \caption{\textbf{Robust object detection by design.} We constrain the Lipschitz
    constant of an SSD detector, to improve its robustness to input perturbations. Each column shows the same image clean (left) and
    under a TOG-Vanishing adversarial attack (right, $\ell_2$ budget
    $\varepsilon = 2$). The rows compare a vanilla SSD (top) with our LipSSD
    (bottom). Without ever training on adversarial examples, LipSSD keeps
    detecting objects that the attack erases for the vanilla model.}
    \label{fig:teaser}
\end{figure}

Object detection is a core component of perception systems, with applications in an
increasing number of fields such as autonomous driving, railways, aeronautics,
healthcare, and robotics. In these settings, instability of the detection system may
lead to catastrophic consequences. Therefore, it is crucial to design object detection
solutions that are not only accurate on clean data, but also stable under input
perturbations. One way to study this stability is to evaluate robustness under
adversarial attacks, which are worst-case perturbations added to the input to cause the
model to fail~\cite{szegedy2014,Goodfellow2014ExplainingAH}. Such attacks provide a
useful stress test for the sensitivity of detection systems to harmful input variations.

The adversarial robustness of neural networks has been extensively studied in the
context of image classification, first through the design of adversarial attacks that
expose their vulnerability to small input
perturbations~\cite{Goodfellow2014ExplainingAH,carlini2017towards,madry2018}, and then
through defense mechanisms such as  adversarial
training~\cite{madry2018,zhang2019trades, wang2019, wang2024} and certified
defenses~\cite{cohen2019,zhang2018,xu2020a,wang2021}, including Lipschitz-constrained
networks~\cite{hu2025,serrurier2021,boissin2025}. Several surveys exist in this domain,
such as~\cite{wang2023adversarial,nguyen2025, zhao2025revisiting}. The literature on adversarial robustness for object
detection is more recent and still very active, with several attack
methods~\cite{xie2017, wei2018transferable, li2018robust, yahn2025adversarial,
nguyen2025}. The object detection task is indeed more complex than classification, as it
adds a localization objective to the classification one, involves multi-scale feature
maps, variable-size output sets with multiple anchors per location, and non-trivial
post-processing steps such as non-maximum suppression (NMS) to produce the final
detections.

Existing work on adversarial defenses for object detection is largely dominated by
adversarial training approaches~\cite{zhang2019,chen2021,chenCWAT2021,dong2022}, which
all rely on providing adversarial examples during training. Other defenses introduce
specialized mechanisms, such as adversarially aware convolutions~\cite{dong2022} or
adversarial patch detection and removal modules~\cite{liu2022segment}. However, these
defenses augment the inference pipeline with dedicated defense components, and their
reliance on specific attacks or objectives can limit transfer to new attacks and
architectures.

In contrast, Lipschitz-constrained architectures aim to improve robustness
(\cref{fig:teaser}) by design. They are not tied to a specific attack model, as they
control the sensitivity of the model to input perturbations by constraining the
Lipschitz constant of the network. Several works have explored Lipschitz-constrained
architectures and their properties in image
classification~\cite{anil2019,li2019,serrurier2021,bethune2022,hu2025}, and recent works
have started to extend them to other tasks such as semantic
segmentation~\cite{massena2025fast}. However, to the best of our knowledge,
\cite{becktor2020} remains the only work to investigate Lipschitz-constrained
architectures for object detection, on a single use-case.

In this work, we explore the design of Lipschitz-constrained detectors and evaluate
their robustness against adversarial attacks. We focus mainly on the Single Shot
MultiBox Detector (SSD) architecture~\cite{liu2016}, which is a popular one-stage
detection model. We propose a Lipschitz-SSD model (LipSSD) which is a
Lipschitz-constrained variant of the original SSD and show that such architectural
Lipschitz control improves adversarial robustness.

Our contributions are as follows:
\begin{itemize}
\item We introduce a robust-by-design Lipschitz-constrained detection architecture which
is attack-agnostic and does not rely on adversarial training.

\item We study the accuracy-robustness trade-off induced by Lipschitz constraints, and
show that it can be controlled through a single training hyperparameter.

\item We perform empirical evaluations under multiple white-box adversarial attacks on
the standard Pascal VOC dataset, as well as on more specific safety-critical datasets
such as LARD~\cite{bougacha2026lard} and KITTI~\cite{geiger2013vision} on which
robustness is improved without sacrificing clean performance.

\item We show that Lipschitz-constrained detectors can be combined with adversarial
training, leading to better adversarial robustness against unseen attacks than the
standard adversarially trained SSD.

\end{itemize}

\vspace{1em}
Code to train and evaluate our Lipschitz-constrained detectors will be made publicly
available.
\section{Background and related works}
\label{sec:background}

In this section, we introduce some background on adversarial attacks for object
detection and the main categories of defenses proposed in the literature. We then focus
on Lipschitz-constrained networks, which provide the foundation of our approach.
\subsection{Adversarial attacks for object detection}
\label{subsec:background_attacks}

Deep neural networks are known to be vulnerable to well-crafted small perturbations of
the input, commonly referred to as adversarial attacks~\cite{szegedy2014}. The
literature on adversarial attacks first focused on classification
tasks~\cite{Goodfellow2014ExplainingAH, carlini2017towards, madry2018}, and was later
extended to other computer vision tasks such as semantic segmentation~\cite{xie2017} and
object detection~\cite{xie2017,chow2020,liu2018dpatch, chen2024overload}. Given an image
$x \in [0,1]^{C \times H \times W}$ and a detector $D_\theta$, an adversarial example
$x_{\mathrm{adv}} = x + \delta^\star$ is obtained by searching for a perturbation
$\delta^\star$ that maximizes an attack objective $\mathcal{L}_{\mathrm{atk}}$ over a
set of admissible perturbations $\Delta_p(\varepsilon)$:
\begin{equation}
\label{eq:adv}
\delta^\star \in
\arg\max_{\delta \in \Delta_p(\varepsilon)}
\mathcal{L}_{\mathrm{atk}}\big(D_\theta(x+\delta), y\big),
\end{equation}
where $\Delta_p(\varepsilon) = \{\, \delta : \|\delta\|_p \leq \varepsilon,\; x+\delta
\in [0,1]^{C \times H \times W} \,\}$. Here, $\mathcal{L}_{\mathrm{atk}}$ can be
designed to target different components of the detection pipeline, including
classification, localization, objectness, or even post-processing steps such as
non-maximum suppression.

Attacks on object detectors may take different forms, including norm-bounded
perturbations, adversarial patches, universal perturbations, and input-specific
perturbations, in either targeted or untargeted settings. \citet{nguyen2025} provide a
recent review of adversarial attacks for object detection. The adversarial examples
crafted by these attacks can fool detectors in multiple ways, for instance by
suppressing correct bounding boxes, creating false detections, modifying predicted
labels, or increasing inference time. \citet{chow2020} provide a taxonomy of these
attacks in a specific framework called TOG, naming the three first as vanishing,
fabrication and mislabelling attacks.

\subsection{Adversarial defenses}
\label{subsec:background_defenses}

Since adversarial attacks expose the sensitivity of object detectors, many works
evaluate robustness by measuring the ability of models to withstand such attacks, in
addition to their standard performance on clean data. Defenses are commonly divided into
empirical defenses and certified defenses. We focus here on empirical defenses, as
certified defenses for object detection are still at an early stage and lie outside the
scope of this work.

\paragraph{Empirical defenses:} Empirical defenses aim to make the model more robust
without providing formal guarantees, and cover several directions in object detection.
The most common one is adversarial training, which uses adversarial examples during
training to improve robustness~\cite{zhang2019,chenCWAT2021,choi2022}. Closely related
approaches, such as RobustDet~\cite{dong2022}, introduce adversarially aware
architectural components to reduce the conflict between clean and adversarial features
during robust training. Similarly,~\citet{liu2022segment} perform adversarial patch
removal using a segmentation module trained with adversarial examples. Other approaches
explore different mechanisms such as ensemble methods~\cite{wu2023} or Gabor
filters~\cite{amirkhani2022adversarial}. The recent work of~\citet{thunuguntla2025}
presents a review of defenses for object detection.

\paragraph{Adversarial training:} Since adversarial training is studied in our
experiments, we describe it in more detail. It trains the model on adversarial examples
generated during optimization and, in its standard form, can be written as the following
min-max optimization problem:
\begin{equation}
\min_\theta \;
\mathbb{E}_{(x,y)\sim\mathcal{D}}
\left[
\mathcal{L}_{\mathrm{det}}
\big(D_\theta(x+\delta^\star), y\big)
\right],
\end{equation}
where $\mathcal{L}_{\mathrm{det}}$ denotes the detection loss and $\delta^\star$ is the
solution to the inner maximization problem described in \cref{eq:adv}. Several variants
of adversarial training have been proposed~\cite{chen2021,chenCWAT2021}, for example to
reduce the training overhead~\cite{shafahi2019}. In object detection, adversarial
training methods usually depend on a specific attack objective, perturbation budget, and
detection loss, which may limit their transfer to unseen attacks or architectures.

\subsection{Lipschitz-constrained neural networks}
\label{subsec:background_lipschitz}

Lipschitz-constrained networks aim to control the worst-case sensitivity of the model to
input perturbations by constraining its Lipschitz constant. More precisely, a
$L$-Lipschitz network $g_\theta$ in the $\ell_{2}$ norm satisfies
\begin{equation}
\label{eq:comp_lip}
\|g_\theta(x+\delta)-g_\theta(x)\|_2
\leq L \|\delta\|_2,
\end{equation}
which provides a direct way to control the worst-case variation of the network output
under bounded input perturbations.

Computing the exact Lipschitz constant of a deep network is
intractable~\cite{virmaux2018lipschitz}. A common strategy is to control it by
construction using the sub-multiplicative property under composition: for a network
$g_\theta = f_n \circ \cdots \circ f_1$ whose layers $f_i$ are each $l_i$-Lipschitz,
\begin{equation}
\mathrm{Lip}(g_\theta)
= \mathrm{Lip}\!\left(f_n \circ \cdots \circ f_1\right)
\leq \prod_{i=1}^{n} \mathrm{Lip}(f_i)
= \prod_{i=1}^{n} l_i,
\end{equation}
so constraining every layer to be $1$-Lipschitz yields a globally $1$-Lipschitz network.
For a linear or convolutional layer $x \mapsto Wx+b$, the $\ell_2$ Lipschitz constant is
equal to the spectral norm $\|W\|_2$, i.e., the largest singular value of $W$. Bounding
the spectral norm through spectral normalization~\cite{miyato2018}, together with
$1$-Lipschitz activation functions, therefore provides a simple way to build
$1$-Lipschitz networks.

\citet{anil2019} observed, however, that only bounding the spectral norm of each layer
leads to gradient-norm attenuation, which limits the expressivity of the Lipschitz
models. They introduced \emph{gradient-norm-preserving} (GNP) layers and replaced
norm-bounded weight matrices with semi-orthogonal ones, satisfying $W^\top W = I$ or $W
W^\top = I$ with $I$ the identity matrix. These constraints can be enforced using
orthogonal parametrizations such as Björck
orthonormalization~\cite{bjorck1971iterative}. Both spectral normalization and these
orthogonal parametrizations enforce the constraint by \emph{reparametrization}: the
trainable weights are deterministically mapped to a constraint-satisfying matrix at
every forward pass, so the Lipschitz bound holds by construction, rather than through a
regularization penalty or Riemannian optimization on the constraint manifold. The
remaining components can also be replaced by norm-preserving counterparts: activation
functions such as GroupSort~\cite{anil2019} instead of ReLU, and $\ell_2$-norm pooling
instead of max-pooling~\cite{boureau2010,serrurier2021}.

Lipschitz-constrained networks are mostly studied for image
classification~\cite{serrurier2021,bethune2022,hu2025}, often motivated by certified
robustness. Indeed, Lipschitz constraints can provide formal guarantees on the
robustness of classifiers. The work of  \citet{massena2025fast} recently extended
Lipschitz-constrained architectures to provide certificates for the semantic
segmentation task.

\paragraph{Lipschitz networks in object detection}
Such guarantees are difficult to obtain in object detection due to IoU matching, NMS,
and bounding-box regression. To the best of our knowledge, ~\citet{becktor2020} are the
only ones to study Lipschitz-constrained architectures for object detection. They
propose to perform object detection using Lipschitz layers without providing any
certificates. To build Lipschitz layers, they apply spectral normalization and
GroupSort~\cite{anil2019} activation functions. However, they only consider a simplified
two-class setup on a custom maritime dataset, and evaluate robustness against common
corruptions (Gaussian noise and blur) rather than adversarial attacks.

Our work follows this direction and studies a Lipschitz detector built with
orthonormalized layers, evaluated against adversarial perturbations rather than common
corruptions, as a means of improving the empirical robustness of object detectors in a
way that is complementary to adversarial training.
\section{Method}
\label{sec:method}

\begin{figure*}
    \centering
    \resizebox{\textwidth}{!}{%
        \input{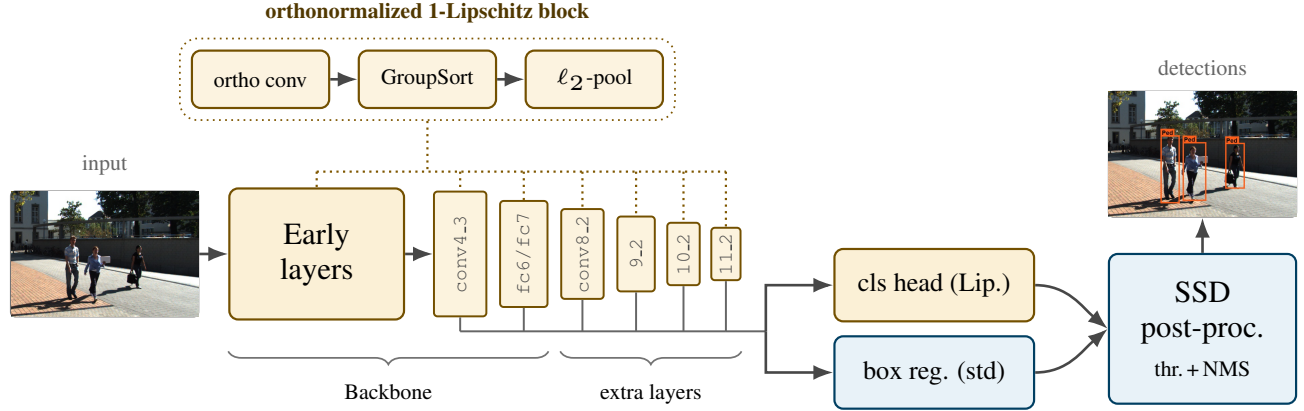}%
    } \caption{\textbf{LipSSD.} Backbone and SSD extra layers, rebuilt from
    $1$-Lipschitz operators (spectral and orthogonal convolutions, GroupSort, $\ell_2$
    pooling), produce six multi-scale maps (\texttt{conv4\_3}--\texttt{conv11\_2}). The
    shared classification head is Lipschitz while the box-regression head stays
    unconstrained. SSD score thresholding and NMS are also unchanged. Brown:
    Lipschitz-constrained, blue: unconstrained box head and standard SSD.}
    \label{fig:lipschitz-ssd-compact}
\end{figure*}

\subsection{Detector architecture selection}

Object detection is an important problem in the field of computer vision that received a
lot of attention. After the image classification task, numerous researchers studied this
more complex task, proposed methods, datasets and challenges. In the last decade,
multiple deep architectures have been proposed including two-stage detectors such as
Faster-RCNN~\cite{ren2015faster} and one-stage detectors such as SSD~\cite{liu2016},
YOLO~\cite{redmon2016you,terven2023comprehensive}, or FCOS~\cite{tian2019}. More
recently transformer-based architectures are addressing object detection, with
architectures such as DETR~\cite{carion2020end}.

We studied the details of multiple detection architecture with the objectives to
constrain the Lipschitz constant of each operation (\cref{eq:comp_lip}). Thus, we do not
consider transformer-based detectors, as constraining the Lipschitz constant of the
self-attention modules is still an open problem. The functions of a convolutional-based
classification architecture can however be constrained and are the backbones of most
convolutional detectors. Unfortunately, the orthogonality constraints on backbones
prevent us from initializing models with  pretrained weights, which may hurt clean
performance. Moreover, proposal and ROI-pooling of two-stage
detectors~\cite{ren2015faster} or the CSP/C2f modules of YOLO
models~\cite{terven2023comprehensive} are harder to control. We thus select the Single
Shot MultiBox Detector (SSD)~\cite{liu2016}, which is commonly used to evaluate
adversarial robustness in object detection~\cite{zhang2019, chenCWAT2021, dong2022,
cheng2025adversarial}.

The SSD model consists of a backbone convolutional network and a set of extra layers,
which produce multi-scale feature maps. These feature maps are then fed into two shared
heads, one for classification and one for bounding box regression. The predictions of
both heads are then aggregated, thresholded based on confidence scores, and filtered
using Non-Maximum Suppression (NMS).

Additional works on the FCOS model are presented in~\cref{app:lipfcos}.

\subsection{LipSSD architecture}

To build our Lipschitz-constrained SSD (LipSSD), we replace every convolutional block by
an orthonormalized $1$-Lipschitz block, as recalled in
\cref{subsec:background_lipschitz}. Bounding only the largest singular
value~\cite{miyato2018} already keeps the layer $1$-Lipschitz, but lets the smaller
singular values attenuate gradients. We instead keep the whole singular-value spectrum
close to $1$, which limits this gradient attenuation~\cite{anil2019}. Then we apply this
orthonormalized $1$-Lipschitz block, depicted in \cref{fig:lipschitz-ssd-compact}, at
every feature map of the network.

\subsubsection{Orthogonal convolutions}

For a convolution kernel $\Phi \in \mathbb{R}^{c_{\mathrm{out}} \times c_{\mathrm{in}}
\times k \times k}$, enforcing $W_\Phi^\top W_\Phi = I$ is not straightforward: $W_\Phi$
is a very large structured, composed of doubly block-Toeplitz, operator, so constraining
this directly is computationally intractable~\cite{boissin2025}. We instead parameterize
the kernel so that the induced operator is orthogonal. For every standard convolution we
use reshaped kernel orthogonalization (RKO)~\cite{serrurier2021}: This reshapes the
kernel into $M \in \mathbb{R}^{c_{\mathrm{out}} \times k^2 c_{\mathrm{in}}}$, rescales
it by its spectral norm through power iteration, and orthonormalizes $M$ with Björck
iterations~\cite{bjorck1971iterative},
\begin{equation}
  M_{t+1} = \tfrac{3}{2} M_t - \tfrac{1}{2}\, M_t M_t^\top M_t ,
  \label{eq:bjorck}
\end{equation}
after which a coercive factor makes the layer $1$-Lipschitz.

RKO does not, however, support the dilated convolution we need for the \texttt{fc6}
layer (\cref{fig:lipschitz-ssd-compact}). There we rely on adaptive orthogonal
convolution (AOC)~\cite{boissin2025}, which composes an RKO factor with an exactly
orthogonal BCOP factor~\cite{li2019} through the operator \textit{block
convolution}~$\circledast$,
\begin{equation}
  \Phi_{\mathrm{AOC}} = \Phi_{\mathrm{RKO}} \circledast \Phi_{\mathrm{BCOP}} ,
  \label{eq:aoc}
\end{equation}
pairing a $(k{+}1{-}s)\times(k{+}1{-}s)$ stride-$1$ BCOP factor with an $s\times s$
stride-$s$ RKO factor. The BCOP factor builds a $k \times k$ orthogonal kernel from
Björck-orthonormal projectors and is itself limited to unit stride. The composition
lifts that limitation, so AOC natively supports dilated, grouped, and transposed
convolutions~\cite{boissin2025}.

\subsubsection{Activation and pooling}

Beyond convolutions, the remaining layers must also be made $1$-Lipschitz. Although ReLU
is 1-Lipschitz, \citet{anil2019} showed that using gradient-norm-preserving activations
such as GroupSort can improve expressiveness, accuracy, and robustness. Accordingly, we
replace ReLU with GroupSort. For pooling layers we also use the $\ell_2$-norm
pooling~\cite{boureau2010}, defined on a window $R_j$ as $\operatorname{pool}(x)_j =
\sqrt{\sum_{i \in R_j} x_i^2}$. This pooling preserves gradient norms only when its
windows do not overlap, i.e.\ when the stride equals the kernel size. We therefore leave
the single overlapping MaxPool of the VGG backbone (kernel size $3$, stride $1$, between
\texttt{conv5\_3} and \texttt{fc6}) unchanged, both for fidelity to the original SSD and
because $\ell_2$-norm pooling does not apply there. This layer is $\sqrt{9} =
3$-Lipschitz~\cite{avant2023analytical}, since, for an overlapping $3 \times 3$
stride-$1$ max-pool, each input feeds at most $9$ output windows. Thus, this only
rescales the global Lipschitz bound of the LipSSD network.

\subsubsection{Assembling LipSSD}

We deploy these building blocks with two complementary libraries. For standard
convolutions, GroupSort activation, and $\ell_2$-norm pooling, we use the
implementations provided by the open-source TorchLip library~\cite{serrurier2021}, while
the dilated \texttt{fc6} layer uses the AOC implementation from
Orthogonium~\cite{boissin2025}, which provides native support for dilation. These
orthonormalized convolutions set LipSSD apart from prior Lipschitz detectors that rely
on plain spectral normalization~\cite{becktor2020}.

Finally, the expressivity of Lipschitz networks for regression remains an open question.
We therefore keep the classification head Lipschitz but leave the box-regression head of
LipSSD unconstrained. \cref{fig:lipschitz-ssd-compact} gives an overview of the
resulting architecture.

\subsubsection{SLipSSD: The RMS scaled version}
\label{sec:slipssd}
As the Pascal VOC dataset is particularly challenging, we study an alternative scaled
version of our LipSSD, which we call \emph{SLipSSD}. This version uses the
$\sqrt{\frac{fan-out}{fan-in}}$ scaling factor proposed first
by~\citet{yang2023spectral}, and then by~\citet{massena2025fast} on convolutions for
Lipschitz networks. This factor is directly multiplied to the output of the
convolutional layers, which allows us to reach better performance at the cost of
increasing the Lipschitz constant of the network.

\subsection{Temperature-scaled classification loss}

For classification tasks, \citet{bethune2022} showed that with Lipschitz neural
networks, the accuracy-robustness trade-off can be controlled through an extra scaling
parameter of the loss function. For example, the cross-entropy (CE) loss can be used
with a temperature parameter $\tau > 0$, as follows: 

\begin{equation}
  \mathcal{L}_{\mathrm{CE}}^\tau(f(x), y)
  = \frac{1}{\tau}\, \mathrm{CE}\!\left(\tau f(x), y\right) .
  \label{eq:tempce}
\end{equation}

The parameter $\tau$ directly controls the classification margins and the sharpness of
the decision boundary. Increasing $\tau$ improves accuracy at the cost of robustness,
while smaller values yield less accurate but more robust networks.

The original SSD loss is composed of a classification term using cross-entropy and a
box-regression term using smooth-$\ell_1$ loss. We apply the temperature scaling only to
the classification CE, while leaving the regression part unchanged. This still allows us
to control the accuracy-robustness trade-off of the detector as we will show
in~\cref{subsubsec:ar_tradeoff}.
\section{Experiments}

\subsection{Robustness evaluation setup}

To evaluate the robustness of our models we use various white-box adversarial attacks:
PGD~\cite{zhang2019} on the classification and localization losses, referenced as
$A_{\mathrm{cls}}$ and $A_{\mathrm{reg}}$ respectively, as in~\cite{chenCWAT2021}, and
DAG~\cite{xie2017}. We also evaluate our models under the targeted attack setup
introduced by~\citet{chow2020}, which includes three attack types: Fabrication (TOG-F),
Vanishing (TOG-V), and Mislabeling (TOG-M). We use a strong attack setting with 40
iterations for PGD, 100 iterations for DAG and 50 iterations for TOG. The number of
iterations is chosen to ensure that the budget of the attack is reached for all models.
We use the standard mAP@50 on clean images and under attack as the evaluation metric. As
our Lipschitz constraint is designed on the $\ell_2$-norm, we evaluate the robustness of
our models in this setup, with different budgets $\varepsilon$ for each dataset. 

\subsection{Benchmark evaluation: Pascal VOC}

\begin{table*}
\centering
\begin{tabular}{llcccccc}
\toprule
Model & Clean & $A_{\mathrm{cls}}$ & $A_{\mathrm{reg}}$ & DAG & TOG-F & TOG-V & TOG-M \\
\midrule
SSD         & \textbf{78.7} & 1.8 & 2.6 & 0.0 & 1.9 & 2.2 & 1.2 \\
SLipSSD & 65.1 & 7.4 & 16.4 & 7.0 & 24.1 & 19.5 & 40.0\\
LipSSD & 56.6 & \textbf{16.6} & \textbf{33.2} & \textbf{25.7} & \textbf{41.4} &
\textbf{37.4} & \textbf{52.0} \\
\bottomrule
\end{tabular}
\caption{mAP@50 for different models under several attacks on the Pascal VOC test set with
attack budget $\varepsilon = 3.0$. $A_{\mathrm{cls}}$ and $A_{\mathrm{reg}}$ denote PGD
targeting the classification and localization losses, respectively. TOG-F, TOG-V, and
TOG-M denote the TOG-Fabrication, TOG-Vanishing, and TOG-Mislabeling attacks,
respectively.}
\label{tab:map50_results_voc_attacks_eps3}
\end{table*}

We use the Pascal VOC~\cite{everingham2015pascal} dataset as a classic benchmark for
robustness in object detection, allowing us to evaluate our approach in a
well-established setting~\cite{zhang2019, chenCWAT2021, dong2022, cheng2025adversarial}.
We adopt the standard “07+12” protocol for training~\cite{dong2022, chenCWAT2021,
zhang2019}, containing $\sim$16k images of 20 categories. For testing, the PASCAL VOC
2007 test set with 4,952 test images is used. 

\textbf{Training details: } We first train a standard SSD with a VGG16 backbone, referred to as SSD in the following, in which the batch normalization layers are removed, as in~\cite{benz2021batch, dong2022}. Training is achieved using the Adam
optimizer~\cite{kingma2014adam} with a learning rate of $10^{-4}$, and a weight decay
of $5\cdot10^{-4}$. We train the two Lipschitz variants, denoted LipSSD and SLipSSD (\cref{sec:slipssd}), with a learning rate of  
$2\cdot10^{-4}$ and no weight decay.
We use
the standard SSD300 data augmentation~\cite{liu2016} pipeline, and resize the images to $300\times300$. 

\begin{figure}[ht]
    \centering
\definecolor{ssdblue}{HTML}{0072B2}
\definecolor{liporange}{HTML}{E69F00}
\definecolor{sliporange}{HTML}{D55E00}
\begin{tikzpicture}
  \begin{axis}[
      width=\linewidth,
      height=0.66\linewidth,
      xlabel={Attack budget $\varepsilon$},
      ylabel={mAP@50 (\%)},
      xmin=-0.08, xmax=3.08,
      ymin=0, ymax=92,
      xtick={0,0.2,0.5,1,2,3},
      ytick={0,20,40,60,80},
      grid=major,
      major grid style={draw=gray!60, dotted},
      tick align=outside,
      axis x line*=bottom,
      axis y line*=left,
      xtick pos=left,
      ytick pos=left,
      legend pos=north east,
      legend cell align={left},
      legend style={draw=gray!40, font=\small, fill=white,
                    fill opacity=0.9, text opacity=1},
      label style={font=\small},
      tick label style={font=\small},
      every axis plot/.append style={line width=1.1pt, mark size=2.4pt},
  ]
    \addplot[ssdblue, mark=*]
      coordinates {(0,78.65) (0.2,64.88) (0.5,39.48) (1,16.88) (2,5.09) (3,2.60)};
    \addlegendentry{SSD}
    \addplot[sliporange, mark=triangle*]
      coordinates {(0,65.06) (0.2,62.75) (0.5,58.02) (1,48.41) (2,28.88) (3,16.50)};
    \addlegendentry{SLipSSD}
    \addplot[liporange, mark=square*]
      coordinates {(0,56.57) (0.2,55.61) (0.5,54.05) (1,50.81) (2,42.45) (3,33.20)};
    \addlegendentry{LipSSD}
  \end{axis}
\end{tikzpicture}
    \caption{mAP@50 on the Pascal VOC test set as the budget $\varepsilon$ of the
    $A_{\mathrm{reg}}$ attack increases. Both Lipschitz variants are substantially more resilient than the standard SSD: 
    despite their lower clean mAP@50, LipSSD and SLipSSD preserve considerably higher performance under attack, even at small perturbation budgets.}
    \label{fig:voc_map50_vs_eps}
\end{figure}
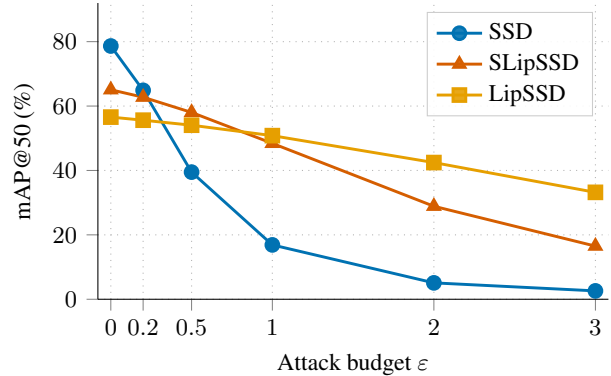

\begin{figure*}[ht]
    \centering
    \input{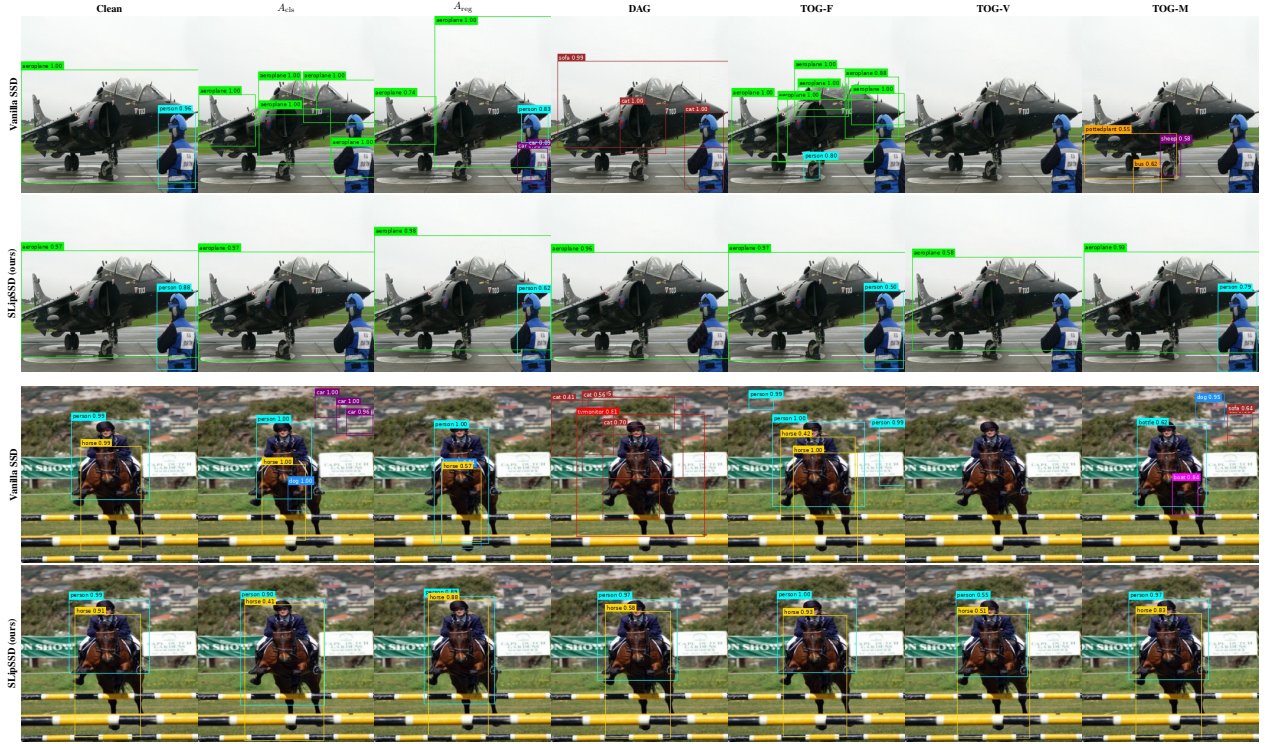}
    \caption{\textbf{SLipSSD preserves detection under attack.} Qualitative comparison
    on two Pascal VOC test images. For the vanilla SSD and our SLipSSD, we show the detections on the clean image and
    under the six adversarial attacks used for evaluation. SLipSSD is more robust.}
    \label{fig:voc_attacks_qualitative}
\end{figure*}

\textbf{Results: } We evaluate the models with a budget of $\varepsilon = 3.0$, see~\cref{tab:map50_results_voc_attacks_eps3}. The SLipSSD is able
to reach better clean performance than the LipSSD, while still improving robustness
compared to the standard SSD. The LipSSD version achieves the best robustness results,
at the cost of a larger drop in clean performance. \Cref{fig:voc_map50_vs_eps} shows the
evaluation of the mAP@50 under the $A_{\mathrm{reg}}$ attack with increasing budget.
Even though the Lipschitz versions degrade clean performance, they show better
performance under attack even for small budgets. \Cref{fig:voc_attacks_qualitative}
shows visual results of the standard SSD and our SLipSSD under the six attacks reported
in the tables. The SLipSSD is able to mostly preserve the detections under attack,
whereas the standard SSD is easily disrupted.

\subsubsection{Accuracy-robustness trade-off}
\label{subsubsec:ar_tradeoff}

\begin{figure}[ht]
    \centering
\begin{tikzpicture}
  \begin{axis}[
      width=0.82\linewidth,
      height=0.6\linewidth,
      xlabel={mAP@50 under attack (\%)},
      ylabel={Clean mAP@50 (\%)},
      xmin=6, xmax=39,
      ymin=48, ymax=70,
      grid=major,
      major grid style={draw=gray!60, dotted},
      tick align=outside,
      axis x line*=bottom,
      axis y line*=left,
      xtick pos=left,
      ytick pos=left,
      label style={font=\small},
      tick label style={font=\small},
      colormap/viridis,
      colorbar,
      point meta min=-1,
      point meta max=0.69897,
      colorbar style={
          ylabel={Temperature $\tau$},
          ytick={-1,-0.30103,0,0.69897},
          yticklabels={0.1,0.5,1,5},
          width=0.28cm,
          ylabel style={font=\small},
          yticklabel style={font=\small},
      },
      scatter/use mapped color={draw=black, fill=mapped color},
  ]
    \addplot[
        scatter, only marks, scatter src=explicit,
        mark=*, mark size=3.2pt,
    ]
      coordinates {
        (10.44,67.1) [0.69897]
        (19.50,65.1) [0]
        (26.42,61.3) [-0.30103]
        (34.98,51.5) [-1]
      };
  \end{axis}
\end{tikzpicture}
    \caption{Accuracy-robustness trade-off on Pascal VOC for the SLipSSD models trained
    with different temperatures $\tau$. Each point reports clean mAP@50 on full test
    set, vertical axis, against robust mAP@50 under the TOG-Vanishing attack at
    $\varepsilon = 3$, horizontal axis. Lowering $\tau$ tightens the effective Lipschitz
    constraint, trading clean accuracy for robustness.}
    \label{fig:voc_tog_vanishing_tradeoff}
\end{figure}
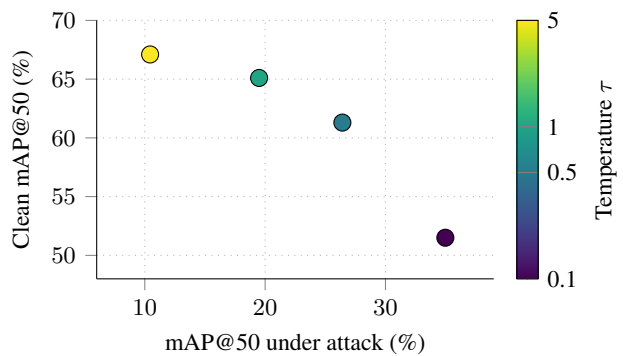

The robustness of neural networks is usually obtained at the cost of a drop in clean
performance (\cref{tab:map50_results_voc_attacks_eps3}), highlighting the trade-off between accuracy and robustness. In our case,
the Lipschitz constant of the network is a hyperparameter that we can manipulate to
target different points of the Pareto front. Following the approach
of~\citet{bethune2022} on the image classification task, we train several versions of
the SLipSSD with different temperature parameters on the Cross-Entropy loss for the
classification component of the global SSD loss. Changing the temperature allows simple
control over the Lipschitz constant of the network and therefore we can reach different
points of the accuracy-robustness trade-off. We report the results
in~\cref{fig:voc_tog_vanishing_tradeoff} for the TOG-Vanishing attack. Note that even if the temperature is only applied on the classification loss, this impacts the whole detector robustness, see~\cref{app:lipssd_quant}.

\subsubsection{Complementarity with adversarial training}

\begin{table*}
\centering
\begin{tabular}{llcccccc}
\toprule
Model & Clean & $A_{\mathrm{cls}}$ & $A_{\mathrm{reg}}$ & DAG &TOG-F & TOG-V & TOG-M \\
\midrule
SSD         & \textbf{78.7} & 0.8 & 1.7 & 0.0 & 0.7 & 0.9 &  0.5  \\
MTD-SSD-BN  & 46.6 & 7.5 & 24.9 & 11.1 &22.5 & 21.9 & 30.3 \\
MTD-SSD     & 64.1 & \textbf{16.8} & \textbf{33.3} & 15.1 & 32.6 & 33.1 & 38.4  \\
\midrule
SLipSSD & 65.1 & 2.7 & 7.2 & 0.7 & 8.3 & 7.3 & 13.6 \\
SLipSSD-MTD & 61.2 & 13.2 & 31.9 & \textbf{20.3} & \textbf{36.2} & \textbf{34.7} &
\textbf{53.7}  \\
\bottomrule
\end{tabular}
\caption{Adversarially trained models results on the Pascal VOC test set with attack budget $\varepsilon = 5.0$.}
\label{tab:map50_results_mtd_eps5}
\end{table*}

To study the complementarity of our method with adversarial training, we implement the
work of~\citet{zhang2019} on both the standard SSD and the proposed SLipSSD. The original work
was conducted under a $\ell_\infty$ setup on a SSD with VGG16 backbone using batch normalization on
the Pascal VOC dataset. Out of fidelity to the original work we train a
\emph{MTD-SSD-BN} following the algorithm proposed and training procedure, where the
adversarial examples are generated using $A_{\mathrm{cls}}$ and $A_{\mathrm{reg}}$ in
the $\ell_2$ setup. Using $\varepsilon = 1$ we reach $46.6$ clean mAP@50. To have a fair
comparison between the SSD and our SLipSSD, we use the Adam optimizer instead of SGD for
both models, and provide also results on a variant without batch normalization in the
\emph{MTD-SSD}. We then train our SLipSSD with the MTD adversarial training, to evaluate the complementarity of the two approaches. The results are reported in~\cref{tab:map50_results_mtd_eps5}. As those
models are more robust than their standard counterparts, we use a higher attack budget
of $\varepsilon = 5.0$. First, we observe that the version without batch normalization, \emph{MTD-SSD}, achieves better results than the original \emph{MTD-SSD-BN} in terms of both robustness and clean performance.
Interestingly, the combination of the MTD algorithm with our SLipSSD produces moderately
worse results than MTD-SSD on the $A_{\mathrm{cls}}$ and $A_{\mathrm{reg}}$ attacks,
which are targeted during training, and achieves better robustness on all the other attacks. This suggests that the MTD algorithm is complementary to our approach, increasing model robustness while providing better generalization to unseen attacks.

\subsection{Safety-critical case studies: LARD and KITTI}

We now evaluate our approach with two datasets related to safety-critical tasks where
robustness and stability are significant. The first one is the
LARD~\cite{bougacha2026lard} dataset with $\sim$57k images for train and $\sim$52k
images in test. The goal is to detect runways in images taken from different flight
simulators, with usually one very small object, or a few ones, per image. The second one
is the KITTI~\cite{geiger2013vision} dataset related to the autonomous driving scenario,
using $\sim$5k train and $\sim$2k test images. This dataset is more challenging than
LARD, as it contains three classes (car, cyclist and pedestrian) with multiple objects
of variable sizes, in realistic scenarios.

\textbf{Training details:}  We use the the same setup as the Pascal VOC dataset, except
for learning rates and image size. We use a learning rate of $2\cdot10^{-4}$ for both
models and resize input to a $1024 \times 1024$ resolution to cope with small objects.

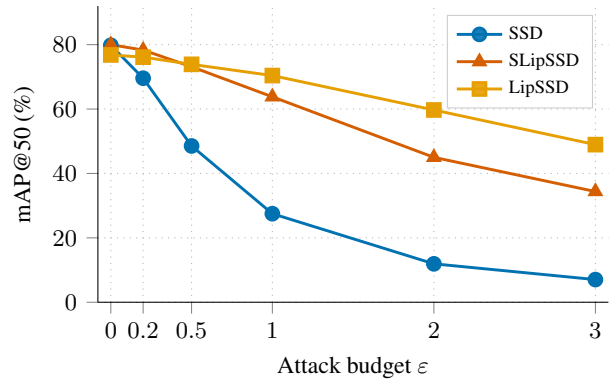
\begin{figure}[ht]
    \centering
\definecolor{ssdblue}{HTML}{0072B2}
\definecolor{liporange}{HTML}{E69F00}
\definecolor{sliporange}{HTML}{D55E00}
\begin{tikzpicture}
  \begin{axis}[
      width=\linewidth,
      height=0.66\linewidth,
      xlabel={Attack budget $\varepsilon$},
      ylabel={mAP@50 (\%)},
      xmin=-0.08, xmax=3.08,
      ymin=0, ymax=92,
      xtick={0,0.2,0.5,1,2,3},
      ytick={0,20,40,60,80},
      grid=major,
      major grid style={draw=gray!60, dotted},
      tick align=outside,
      axis x line*=bottom,
      axis y line*=left,
      xtick pos=left,
      ytick pos=left,
      legend pos=north east,
      legend cell align={left},
      legend style={draw=gray!40, font=\scriptsize, fill=white,
                    fill opacity=0.9, text opacity=1},
      label style={font=\small},
      tick label style={font=\small},
      every axis plot/.append style={line width=1.1pt, mark size=2.4pt},
  ]
    \addplot[ssdblue, mark=*]
      coordinates {(0,79.80) (0.2,69.58) (0.5,48.54) (1,27.50) (2,11.93) (3,7.04)};
    \addlegendentry{SSD}
    \addplot[sliporange, mark=triangle*]
      coordinates {(0,80.00) (0.2,78.32) (0.5,73.13) (1,63.80) (2,45.00) (3,34.38)};
    \addlegendentry{SLipSSD}
    \addplot[liporange, mark=square*]
      coordinates {(0,76.80) (0.2,76.12) (0.5,73.88) (1,70.40) (2,59.71) (3,48.95)};
    \addlegendentry{LipSSD}
  \end{axis}
\end{tikzpicture}
    \caption{mAP@50 on the KITTI test set as the budget $\varepsilon$ of the
  $A_{\mathrm{reg}}$ attack increases. LipSSD and SLipSSD maintain higher robustness
  than the standard SSD over the full budget range.}
    \label{fig:kitti_map50_vs_eps}
\end{figure}

\textbf{Results: }This bigger image resolution increases the computational cost of the
attack, making a full evaluation on the ~52k test images on LARD impractical. We
therefore use 1000 samples randomly selected in the test set for the robustness
evaluation of LARD, but clean performance is reported on the full test set. We report
results of our experiments, both on KITTI and LARD, see
\cref{tab:map50_results_safety_critical_attacks}. Our models improve the robustness of
the standard SSD model on both datasets while largely preserving clean performance. On
KITTI, SLipSSD still achieves higher mAP@50 than LipSSD under TOG-V and TOG-M attacks, mainly due to
its stronger clean performance. Under larger attack budgets, however, LipSSD
consistently outperforms SLipSSD across all attacks, as shown in~\cref{app:lipssd_quant}. \Cref{fig:kitti_map50_vs_eps} highlights this gap as a function of the attack
budget on KITTI: under $A_{\mathrm{reg}}$ attack, LipSSD retains $70.4\%$ mAP@50 at
$\varepsilon = 1$ and still $48.9\%$ at $\varepsilon = 3$, whereas the standard SSD
drops to $27.5\%$ and $7.0\%$.

\begin{table*}
\centering
\begin{tabular}{llccccccc}
\toprule
Dataset & Model & Clean & $A_{\mathrm{cls}}$ & $A_{\mathrm{reg}}$ & DAG & TOG-F & TOG-V
& TOG-M \\
\midrule
\multirow{3}{*}{KITTI} & SSD     & 79.8 & 22.9 & 27.5 & 20.9 & 21.1 & 32.7 & 34.8 \\
  & SLipSSD & \textbf{80.0} & 53.8 & 63.8 & 67.2 & 66.5 & \textbf{76.3} & \textbf{76.3}
  \\
  & LipSSD  & 76.8 & \textbf{65.3} & \textbf{70.4} & \textbf{73.6} & \textbf{74.0} &
  75.9 & 76.4 \\
\midrule
\multirow{3}{*}{LARD} & SSD     & 95.1 & 2.6 & 1.6 & N/A & 4.3 & 16.0 & N/A \\
  & SLipSSD & \textbf{95.4} & 17.2 & 15.4 & N/A & 34.8 & 24.4 & N/A \\
  & LipSSD  & 94.8 & \textbf{22.4} & \textbf{20.3} & N/A & \textbf{45.8} & \textbf{47.7}
  & N/A \\
\bottomrule
\end{tabular}
\caption{mAP@50 for SSD and LipSSD under several attacks on the LARD and KITTI test sets
with attack budget $\varepsilon = 1.0$.}
\label{tab:map50_results_safety_critical_attacks}
\end{table*}

\section{Conclusion and perspectives}
\label{sec:conclusion}

In this work, we present LipSSD, a single-shot object detector with Lipschitz-constrained operations
to improve the models adversarial robustness.
We show that LipSSD also allows simple control of the accuracy-robustness trade-off and can be
combined with adversarial training for additional benefits.

\textbf{Limitations and future work.} The main limitation of our approach is that
training Lipschitz-constrained models is more computationally expensive than training
standard models. In our experiments,
training a LipSSD model takes approximately $2\times$ longer than
a standard SSD. However,
there is no inference overhead, as the learned weights can be used in
regular convolutional layers. Additionally, even though LipSSD maintains a competitive
performance on the KITTI and LARD datasets, our model is not yet able to reach the
clean performance of SSD on Pascal VOC. We hypothesize that the robustness task is more
difficult on datasets like Pascal VOC. More complex models, or 1-Lipschitz pretrained
backbones on ImageNet could therefore be required to reach competitive clean performance.

Building such networks could lead future research on certified detectors. Indeed,
certifying Lipschitz networks has been explored for classification and
segmentation tasks, but not yet for object detection. We believe that such constraints
are a promising direction to certify object detectors.

\section{Acknowledgements}
\label{sec:acknowledgements}

Our work has benefited from the AI Cluster ANITI and the research program
DEEL\footnote{\url{https://www.deel.ai/}}. ANITI is funded by the France 2030 program
under the Grant agreement n°ANR-23-IACL-0002. DEEL is an integrative program of the AI
Cluster ANITI, designed and operated jointly with IRT Saint Exupéry, with the financial
support from its industrial and academic partners and the France 2030 program under the
Grant agreement n°ANR-10-AIRT-01.

The authors would like to thank Clément Lefebvre for his valuable feedbacks on the
manuscript, as well as Quentin Possamaï, Arnaud Jaoul, Kenza Saiah and Nenad Mijatovic
for their continuous support during the project.

{
    \small
    \bibliographystyle{ieeenat_fullname}
    \bibliography{references}

\begin{thebibliography}{63}
\providecommand{\natexlab}[1]{#1}
\providecommand{\url}[1]{\texttt{#1}}
\expandafter\ifx\csname urlstyle\endcsname\relax
  \providecommand{\doi}[1]{doi: #1}\else
  \providecommand{\doi}{doi: \begingroup \urlstyle{rm}\Url}\fi

\bibitem[Amirkhani and Karimi(2022)]{amirkhani2022adversarial}
Abdollah Amirkhani and Mohammad~Parsa Karimi.
\newblock Adversarial defenses for object detectors based on gabor
  convolutional layers.
\newblock \emph{The visual computer}, 38\penalty0 (6):\penalty0 1929--1944,
  2022.

\bibitem[Anil et~al.(2019)Anil, Lucas, and Grosse]{anil2019}
Cem Anil, James Lucas, and Roger Grosse.
\newblock Sorting out {L}ipschitz function approximation.
\newblock In \emph{Proceedings of the 36th International Conference on Machine
  Learning}. PMLR, 2019.

\bibitem[Avant and Morgansen(2023)]{avant2023analytical}
Trevor Avant and Kristi~A Morgansen.
\newblock Analytical bounds on the local lipschitz constants of relu networks.
\newblock \emph{IEEE Transactions on Neural Networks and Learning Systems},
  35\penalty0 (10):\penalty0 13902--13913, 2023.

\bibitem[Becktor et~al.(2020)Becktor, Schöller, Boukas, Blanke, and
  Nalpantidis]{becktor2020}
Jonathan Becktor, Frederik Schöller, Evangelos Boukas, Mogens Blanke, and
  Lazaros Nalpantidis.
\newblock Lipschitz {Constrained} {Neural} {Networks} for {Robust} {Object}
  {Detection} at {Sea}.
\newblock \emph{IOP Conference Series: Materials Science and Engineering},
  929\penalty0 (1), 2020.

\bibitem[Benz et~al.(2021)Benz, Zhang, and Kweon]{benz2021batch}
Philipp Benz, Chaoning Zhang, and In~So Kweon.
\newblock Batch normalization increases adversarial vulnerability and decreases
  adversarial transferability: A non-robust feature perspective.
\newblock In \emph{ICCV}, pages 7818--7827, 2021.

\bibitem[B\'{e}thune et~al.(2022)B\'{e}thune, Boissin, Serrurier, Mamalet,
  Friedrich, and Gonzalez~Sanz]{bethune2022}
Louis B\'{e}thune, Thibaut Boissin, Mathieu Serrurier, Franck Mamalet, Corentin
  Friedrich, and Alberto Gonzalez~Sanz.
\newblock Pay attention to your loss : understanding misconceptions about
  {Lipschitz} neural networks.
\newblock In \emph{NeurIPS}, 2022.

\bibitem[Bj{\"o}rck and Bowie(1971)]{bjorck1971iterative}
{\AA}ke Bj{\"o}rck and C. Bowie.
\newblock An iterative algorithm for computing the best estimate of an
  orthogonal matrix.
\newblock \emph{SIAM Journal on Numerical Analysis}, 8\penalty0 (2):\penalty0
  358--364, 1971.

\bibitem[Boissin et~al.(2025)Boissin, Mamalet, Fel, Picard, Massena, and
  Serrurier]{boissin2025}
Thibaut Boissin, Franck Mamalet, Thomas Fel, Agustin~Martin Picard, Thomas
  Massena, and Mathieu Serrurier.
\newblock An adaptive orthogonal convolution scheme for efficient and flexible
  {CNN} architectures.
\newblock In \emph{ICML}, 2025.

\bibitem[Bougacha et~al.(2026)Bougacha, Delhomme, Ducoffe, Fuchs, Ginestet,
  Girard, Kraiem, Mamalet, Mussot, Pagetti, and Sammour]{bougacha2026lard}
Yassine Bougacha, Geoffrey Delhomme, M{\'e}lanie Ducoffe, Augustin Fuchs,
  Jean-Brice Ginestet, Jacques Girard, Sofiane Kraiem, Franck Mamalet, Vincent
  Mussot, Claire Pagetti, and Thierry Sammour.
\newblock Lard 2.0: Enhanced datasets and benchmarking for autonomous landing
  systems.
\newblock In \emph{ERTS}, 2026.

\bibitem[Boureau et~al.(2010)Boureau, Ponce, and LeCun]{boureau2010}
Y-Lan Boureau, Jean Ponce, and Yann LeCun.
\newblock A theoretical analysis of feature pooling in visual recognition.
\newblock In \emph{ICML}, 2010.

\bibitem[Carion et~al.(2020)Carion, Massa, Synnaeve, Usunier, Kirillov, and
  Zagoruyko]{carion2020end}
Nicolas Carion, Francisco Massa, Gabriel Synnaeve, Nicolas Usunier, Alexander
  Kirillov, and Sergey Zagoruyko.
\newblock End-to-end object detection with transformers.
\newblock In \emph{ECCV}, pages 213--229. Springer, 2020.

\bibitem[Carlini and Wagner(2017)]{carlini2017towards}
Nicholas Carlini and David Wagner.
\newblock Towards evaluating the robustness of neural networks.
\newblock In \emph{2017 ieee symposium on security and privacy (sp)}, pages
  39--57. Ieee, 2017.

\bibitem[Chen et~al.(2024)Chen, Chen, Chung, Lee, et~al.]{chen2024overload}
Erh-Chung Chen, Pin-Yu Chen, I Chung, Che-Rung Lee, et~al.
\newblock Overload: Latency attacks on object detection for edge devices.
\newblock In \emph{CVPR}, 2024.

\bibitem[Chen et~al.(2021{\natexlab{a}})Chen, Kung, and Chen]{chenCWAT2021}
Pin-Chun Chen, Bo-Han Kung, and Jun-Cheng Chen.
\newblock Class-{{Aware Robust Adversarial Training}} for {{Object Detection}}.
\newblock In \emph{CVPR}, 2021{\natexlab{a}}.

\bibitem[Chen et~al.(2021{\natexlab{b}})Chen, Xie, Tan, Zhang, Hsieh, and
  Gong]{chen2021}
Xiangning Chen, Cihang Xie, Mingxing Tan, Li Zhang, Cho-Jui Hsieh, and Boqing
  Gong.
\newblock Robust and accurate object detection via adversarial learning.
\newblock In \emph{CVPR}, 2021{\natexlab{b}}.

\bibitem[Cheng et~al.(2025)Cheng, Huang, Fang, Han, and
  Wang]{cheng2025adversarial}
Jikang Cheng, Baojin Huang, Yan Fang, Zhen Han, and Zhongyuan Wang.
\newblock Adversarial intensity awareness for robust object detection.
\newblock \emph{Computer Vision and Image Understanding}, 251:\penalty0 104252,
  2025.

\bibitem[Chiang et~al.(2020)Chiang, Curry, Abdelkader, Kumar, Dickerson, and
  Goldstein]{chiang2020}
Ping-yeh Chiang, Michael Curry, Ahmed Abdelkader, Aounon Kumar, John Dickerson,
  and Tom Goldstein.
\newblock Detection as {Regression}: {Certified} {Object} {Detection} with
  {Median} {Smoothing}.
\newblock In \emph{NeurIPS}. Curran Associates, Inc., 2020.

\bibitem[Choi and Tian(2022)]{choi2022}
Jung~Im Choi and Qing Tian.
\newblock Adversarial {{Attack}} and {{Defense}} of {{YOLO Detectors}} in
  {{Autonomous Driving Scenarios}}.
\newblock In \emph{2022 {{IEEE Intelligent Vehicles Symposium}} ({{IV}})},
  2022.

\bibitem[Chow et~al.(2020)Chow, Liu, Loper, Bae, Gursoy, Truex, Wei, and
  Wu]{chow2020}
Ka-Ho Chow, Ling Liu, Margaret Loper, Juhyun Bae, Mehmet~Emre Gursoy, Stacey
  Truex, Wenqi Wei, and Yanzhao Wu.
\newblock Adversarial objectness gradient attacks in real-time object detection
  systems.
\newblock In \emph{2020 Second IEEE International Conference on Trust, Privacy
  and Security in Intelligent Systems and Applications (TPS-ISA)}, pages
  263--272, 2020.

\bibitem[Cohen et~al.(2019)Cohen, Rosenfeld, and Kolter]{cohen2019}
Jeremy Cohen, Elan Rosenfeld, and Zico Kolter.
\newblock Certified {{Adversarial Robustness}} via {{Randomized Smoothing}}.
\newblock In \emph{ICML}, 2019.

\bibitem[Cohen et~al.(2025)Cohen, Ducoffe, Boumazouza, Gabreau, Pagetti, Pucel,
  and Galametz]{cohen2025verifiou}
No{\'e}mie Cohen, M{\'e}lanie Ducoffe, Ryma Boumazouza, Christophe Gabreau,
  Claire Pagetti, Xavier Pucel, and Audrey Galametz.
\newblock {{VerifIoU}}: Robustness of object detection to perturbations.
\newblock In \emph{44th {{Digital Avionics Systems Conference}} ({{DASC}})},
  pages 1--10, Montreal, Canada, 2025. {IEEE}.

\bibitem[Dong et~al.(2022)Dong, Wei, and Lin]{dong2022}
Ziyi Dong, Pengxu Wei, and Liang Lin.
\newblock Adversarially-{{Aware Robust Object Detector}}.
\newblock In \emph{ECCV}, 2022.

\bibitem[Ducoffe et~al.(2023)Ducoffe, Carrere, F{\'e}liers, Gauffriau, Mussot,
  Pagetti, and Sammour]{ducoffe2023lard}
M{\'e}lanie Ducoffe, Maxime Carrere, L{\'e}o F{\'e}liers, Adrien Gauffriau,
  Vincent Mussot, Claire Pagetti, and Thierry Sammour.
\newblock Lard--landing approach runway detection--dataset for vision based
  landing.
\newblock \emph{arXiv preprint arXiv:2304.09938}, 2023.

\bibitem[Everingham et~al.(2015)Everingham, Eslami, Van~Gool, Williams, Winn,
  and Zisserman]{everingham2015pascal}
Mark Everingham, SM~Ali Eslami, Luc Van~Gool, Christopher~KI Williams, John
  Winn, and Andrew Zisserman.
\newblock The {P}ascal {V}isual {O}bject {C}lasses {C}hallenge: A
  retrospective.
\newblock \emph{IJCV}, 111\penalty0 (1):\penalty0 98--136, 2015.

\bibitem[Geiger et~al.(2013)Geiger, Lenz, Stiller, and
  Urtasun]{geiger2013vision}
Andreas Geiger, Philip Lenz, Christoph Stiller, and Raquel Urtasun.
\newblock Vision meets robotics: The kitti dataset.
\newblock \emph{The international journal of robotics research}, 32\penalty0
  (11):\penalty0 1231--1237, 2013.

\bibitem[Goodfellow et~al.(2014)Goodfellow, Shlens, and
  Szegedy]{Goodfellow2014ExplainingAH}
Ian~J. Goodfellow, Jonathon Shlens, and Christian Szegedy.
\newblock Explaining and harnessing adversarial examples.
\newblock \emph{CoRR}, abs/1412.6572, 2014.

\bibitem[Hu et~al.(2025)Hu, Hu, and Fredrikson]{hu2025}
Kai Hu, Haoqi Hu, and Matt Fredrikson.
\newblock {{LipNeXt}}: {{Scaling}} up {{Lipschitz-based Certified Robustness}}
  to {{Billion-parameter Models}}.
\newblock In \emph{ICLR}, 2025.

\bibitem[Ilyas et~al.(2019)Ilyas, Santurkar, Tsipras, Engstrom, Tran, and
  Madry]{ilyas2019adversarial}
Andrew Ilyas, Shibani Santurkar, Dimitris Tsipras, Logan Engstrom, Brandon
  Tran, and Aleksander Madry.
\newblock Adversarial examples are not bugs, they are features.
\newblock \emph{Advances in neural information processing systems}, 32, 2019.

\bibitem[Kingma and Ba(2014)]{kingma2014adam}
Diederik~P Kingma and Jimmy Ba.
\newblock Adam: A method for stochastic optimization.
\newblock \emph{arXiv preprint arXiv:1412.6980}, 2014.

\bibitem[Li et~al.(2019)Li, Haque, Anil, Lucas, Grosse, and Jacobsen]{li2019}
Qiyang Li, Saminul Haque, Cem Anil, James Lucas, Roger Grosse, and
  J{\"o}rn-Henrik Jacobsen.
\newblock Preventing gradient attenuation in lipschitz constrained
  convolutional networks.
\newblock In \emph{NeurIPS}, 2019.

\bibitem[Li et~al.(2025)Li, Chen, and Hu]{li2025importance}
Xiao Li, Hang Chen, and Xiaolin Hu.
\newblock On the importance of backbone to the adversarial robustness of object
  detectors.
\newblock \emph{IEEE Transactions on Information Forensics and Security}, 2025.

\bibitem[Li et~al.(2018)Li, Tian, Chang, Bian, and Lyu]{li2018robust}
Yuezun Li, Daniel Tian, Ming-Ching Chang, Xiao Bian, and Siwei Lyu.
\newblock Robust adversarial perturbation on deep proposal-based models.
\newblock In \emph{BMVC}, 2018.

\bibitem[Liu et~al.(2022)Liu, Levine, Lau, Chellappa, and
  Feizi]{liu2022segment}
Jiang Liu, Alexander Levine, Chun~Pong Lau, Rama Chellappa, and Soheil Feizi.
\newblock Segment and complete: Defending object detectors against adversarial
  patch attacks with robust patch detection.
\newblock In \emph{CVPR}, pages 14973--14982, 2022.

\bibitem[Liu et~al.(2016)Liu, Anguelov, Erhan, Szegedy, Reed, Fu, and
  Berg]{liu2016}
Wei Liu, Dragomir Anguelov, Dumitru Erhan, Christian Szegedy, Scott Reed,
  Cheng-Yang Fu, and Alexander~C. Berg.
\newblock {{SSD}}: {{Single Shot MultiBox Detector}}.
\newblock In \emph{ECCV}, 2016.

\bibitem[Liu et~al.(2018)Liu, Yang, Liu, Song, Li, and Chen]{liu2018dpatch}
Xin Liu, Huanrui Yang, Ziwei Liu, Linghao Song, Hai Li, and Yiran Chen.
\newblock Dpatch: An adversarial patch attack on object detectors.
\newblock \emph{arXiv preprint arXiv:1806.02299}, 2018.

\bibitem[Madry et~al.(2018)Madry, Makelov, Schmidt, Tsipras, and
  Vladu]{madry2018}
Aleksander Madry, Aleksandar Makelov, Ludwig Schmidt, Dimitris Tsipras, and
  Adrian Vladu.
\newblock Towards {{Deep Learning Models Resistant}} to {{Adversarial
  Attacks}}.
\newblock In \emph{ICLR}, 2018.

\bibitem[Massena et~al.(2025)Massena, Friedrich, Mamalet, and
  Serrurier]{massena2025fast}
Thomas Massena, Corentin Friedrich, Franck Mamalet, and Mathieu Serrurier.
\newblock Fast and flexible robustness certificates for semantic segmentation.
\newblock \emph{arXiv preprint arXiv:2512.06010}, 2025.

\bibitem[Miyato et~al.(2018)Miyato, Kataoka, Koyama, and Yoshida]{miyato2018}
Takeru Miyato, Toshiki Kataoka, Masanori Koyama, and Yuichi Yoshida.
\newblock Spectral {{Normalization}} for {{Generative Adversarial Networks}}.
\newblock In \emph{ICLR}, 2018.

\bibitem[Nguyen et~al.(2025)Nguyen, Zhang, Lu, Wu, Zheng, Li~Tan, and
  Zhen]{nguyen2025}
Khoi Nguyen~Tiet Nguyen, Wenyu Zhang, Kangkang Lu, Yu-Huan Wu, Xingjian Zheng,
  Hui Li~Tan, and Liangli Zhen.
\newblock A {Survey} and {Evaluation} of {Adversarial} {Attacks} in {Object}
  {Detection}.
\newblock \emph{IEEE Transactions on Neural Networks and Learning Systems},
  36\penalty0 (9):\penalty0 15706--15722, 2025.

\bibitem[Nirala and Sarkar(2025)]{nirala2025}
Ashutosh~Kumar Nirala and Soumalya Sarkar.
\newblock Towards {{Certified Object Detectors}}: {{Certified Runway Detection
  Using Yolo}}.
\newblock In \emph{ICIP}, 2025.

\bibitem[Redmon et~al.(2016)Redmon, Divvala, Girshick, and
  Farhadi]{redmon2016you}
Joseph Redmon, Santosh Divvala, Ross Girshick, and Ali Farhadi.
\newblock You only look once: Unified, real-time object detection.
\newblock In \emph{CVPR}, pages 779--788, 2016.

\bibitem[Ren et~al.(2015)Ren, He, Girshick, and Sun]{ren2015faster}
Shaoqing Ren, Kaiming He, Ross Girshick, and Jian Sun.
\newblock Faster r-cnn: Towards real-time object detection with region proposal
  networks.
\newblock In \emph{NeurIPS}, 2015.

\bibitem[Serrurier et~al.(2021)Serrurier, Mamalet, {Gonzalez-Sanz}, Boissin,
  Loubes, and Del~Barrio]{serrurier2021}
Mathieu Serrurier, Franck Mamalet, Alberto {Gonzalez-Sanz}, Thibaut Boissin,
  Jean-Michel Loubes, and Eustasio Del~Barrio.
\newblock Achieving robustness in classification using optimal transport with
  hinge regularization.
\newblock In \emph{CVPR}, 2021.

\bibitem[Shafahi et~al.(2019)Shafahi, Najibi, Ghiasi, Xu, Dickerson, Studer,
  Davis, Taylor, and Goldstein]{shafahi2019}
Ali Shafahi, Mahyar Najibi, Mohammad~Amin Ghiasi, Zheng Xu, John Dickerson,
  Christoph Studer, Larry~S Davis, Gavin Taylor, and Tom Goldstein.
\newblock Adversarial training for free!
\newblock In \emph{NeurIPS}, 2019.

\bibitem[Szegedy et~al.(2014)Szegedy, Zaremba, Sutskever, Bruna, Erhan,
  Goodfellow, and Fergus]{szegedy2014}
Christian Szegedy, Wojciech Zaremba, Ilya Sutskever, Joan Bruna, Dumitru Erhan,
  Ian Goodfellow, and Rob Fergus.
\newblock Intriguing properties of {{Neural Networks}}.
\newblock In \emph{ICLR}, 2014.

\bibitem[Terven et~al.(2023)Terven, C{\'o}rdova-Esparza, and
  Romero-Gonz{\'a}lez]{terven2023comprehensive}
Juan Terven, Diana-Margarita C{\'o}rdova-Esparza, and Julio-Alejandro
  Romero-Gonz{\'a}lez.
\newblock A comprehensive review of {YOLO} architectures in computer vision:
  From {YOLOv1} to {YOLOv8} and {YOLO-NAS}.
\newblock \emph{Machine learning and knowledge extraction}, 5\penalty0
  (4):\penalty0 1680--1716, 2023.

\bibitem[Thunuguntla et~al.(2025)Thunuguntla, Tadepalli, Raffa, Thunuguntla,
  Tadepalli, and Raffa]{thunuguntla2025}
Anant Thunuguntla, Prasad Tadepalli, Giuseppe Raffa, Anant Thunuguntla, Prasad
  Tadepalli, and Giuseppe Raffa.
\newblock Defenses {Against} {Adversarial} {Attacks} on {Object} {Detection}:
  {Methods} and {Future} {Directions}.
\newblock \emph{Information}, 16\penalty0 (11), 2025.

\bibitem[Tian et~al.(2019)Tian, Shen, Chen, and He]{tian2019}
Zhi Tian, Chunhua Shen, Hao Chen, and Tong He.
\newblock {{FCOS}}: {{Fully Convolutional One-Stage Object Detection}}.
\newblock In \emph{ICCV}, 2019.

\bibitem[Virmaux and Scaman(2018)]{virmaux2018lipschitz}
Aladin Virmaux and Kevin Scaman.
\newblock Lipschitz regularity of deep neural networks: analysis and efficient
  estimation.
\newblock In \emph{NeurIPS}, 2018.

\bibitem[Wang et~al.(2021)Wang, Zhang, Xu, Lin, Jana, Hsieh, and
  Kolter]{wang2021}
Shiqi Wang, Huan Zhang, Kaidi Xu, Xue Lin, Suman Jana, Cho-Jui Hsieh, and
  J.~Zico Kolter.
\newblock Beta-{{CROWN}}: {{Efficient Bound Propagation}} with {{Per-neuron
  Split Constraints}} for {{Neural Network Robustness Verification}}.
\newblock In \emph{NeurIPS}, 2021.

\bibitem[Wang et~al.(2019)Wang, Zou, Yi, Bailey, Ma, and Gu]{wang2019}
Yisen Wang, Difan Zou, Jinfeng Yi, James Bailey, Xingjun Ma, and Quanquan Gu.
\newblock Improving {{Adversarial Robustness Requires Revisiting Misclassified
  Examples}}.
\newblock In \emph{ICLR}, 2019.

\bibitem[Wang et~al.(2023)Wang, Sun, Li, Yuan, Ni, Hossain, and
  Poor]{wang2023adversarial}
Yulong Wang, Tong Sun, Shenghong Li, Xin Yuan, Wei Ni, Ekram Hossain, and
  H~Vincent Poor.
\newblock Adversarial attacks and defenses in machine learning-empowered
  communication systems and networks: A contemporary survey.
\newblock \emph{IEEE Communications Surveys \& Tutorials}, 25\penalty0
  (4):\penalty0 2245--2298, 2023.

\bibitem[Wang et~al.(2024)Wang, Li, Zhu, and Xie]{wang2024}
Zeyu Wang, Xianhang Li, Hongru Zhu, and Cihang Xie.
\newblock Revisiting {{Adversarial Training}} at {{Scale}}.
\newblock In \emph{CVPR}, 2024.

\bibitem[Wei et~al.(2018)Wei, Liang, Chen, and Cao]{wei2018transferable}
Xingxing Wei, Siyuan Liang, Ning Chen, and Xiaochun Cao.
\newblock Transferable adversarial attacks for image and video object
  detection.
\newblock \emph{arXiv preprint arXiv:1811.12641}, 2018.

\bibitem[Wu et~al.(2023)Wu, Chow, Wei, and Liu]{wu2023}
Yanzhao Wu, Ka-Ho Chow, Wenqi Wei, and Ling Liu.
\newblock Exploring {{Model Learning Heterogeneity}} for {{Boosting Ensemble
  Robustness}}.
\newblock In \emph{2023 {{IEEE International Conference}} on {{Data Mining}}
  ({{ICDM}})}, 2023.

\bibitem[Xie et~al.(2017)Xie, Wang, Zhang, Zhou, Xie, and Yuille]{xie2017}
Cihang Xie, Jianyu Wang, Zhishuai Zhang, Yuyin Zhou, Lingxi Xie, and Alan
  Yuille.
\newblock Adversarial {{Examples}} for {{Semantic Segmentation}} and {{Object
  Detection}}.
\newblock In \emph{ICCV}, 2017.

\bibitem[Xu et~al.(2020)Xu, Shi, Zhang, Wang, Chang, Huang, Kailkhura, Lin, and
  Hsieh]{xu2020a}
Kaidi Xu, Zhouxing Shi, Huan Zhang, Yihan Wang, Kai-Wei Chang, Minlie Huang,
  Bhavya Kailkhura, Xue Lin, and Cho-Jui Hsieh.
\newblock Automatic {{Perturbation Analysis}} for {{Scalable Certified
  Robustness}} and {{Beyond}}.
\newblock In \emph{NeurIPS}, 2020.

\bibitem[Yahn et~al.(2025)Yahn, Tekin, Ilhan, Hu, Huang, Xu, Loper, and
  Liu]{yahn2025adversarial}
Zachary Yahn, Selim~Furkan Tekin, Fatih Ilhan, Sihao Hu, Tiansheng Huang,
  Yichang Xu, Margaret Loper, and Ling Liu.
\newblock Adversarial attention perturbations for large object detection
  transformers.
\newblock In \emph{ICCV}, 2025.

\bibitem[Yang et~al.(2023)Yang, Simon, and Bernstein]{yang2023spectral}
Greg Yang, James~B Simon, and Jeremy Bernstein.
\newblock A spectral condition for feature learning.
\newblock \emph{arXiv preprint arXiv:2310.17813}, 2023.

\bibitem[Zhang and Wang(2019)]{zhang2019}
Haichao Zhang and Jianyu Wang.
\newblock Towards adversarially robust object detection.
\newblock In \emph{ICCV}, 2019.

\bibitem[Zhang et~al.(2018)Zhang, Weng, Chen, Hsieh, and Daniel]{zhang2018}
Huan Zhang, Tsui-Wei Weng, Pin-Yu Chen, Cho-Jui Hsieh, and Luca Daniel.
\newblock Efficient {{Neural Network Robustness Certification}} with {{General
  Activation Functions}}.
\newblock In \emph{NeurIPS}, 2018.

\bibitem[Zhang et~al.(2019)Zhang, Yu, Jiao, Xing, El~Ghaoui, and
  Jordan]{zhang2019trades}
Hongyang Zhang, Yaodong Yu, Jiantao Jiao, Eric~P. Xing, Laurent El~Ghaoui, and
  Michael~I. Jordan.
\newblock Theoretically principled trade-off between robustness and accuracy.
\newblock In \emph{ICML}, 2019.

\bibitem[Zhao et~al.(2025)Zhao, Zhang, Li, Sicre, Amsaleg, Backes, Li, Wang,
  and Shen]{zhao2025revisiting}
Zhengyu Zhao, Hanwei Zhang, Renjue Li, Ronan Sicre, Laurent Amsaleg, Michael
  Backes, Qi Li, Qian Wang, and Chao Shen.
\newblock Revisiting transferable adversarial images: Systemization,
  evaluation, and new insights.
\newblock \emph{IEEE Transactions on Pattern Analysis and Machine
  Intelligence}, 2025.

\end{thebibliography}
}

\clearpage
\appendix
\onecolumn

\crefalias{section}{appendix}
\crefalias{subsection}{appendix}
\crefalias{subsubsection}{appendix}
\crefname{appendix}{Appendix}{Appendices}
\Crefname{appendix}{Appendix}{Appendices}

\section{LipFCOS}
\label{app:lipfcos}

This section studies the generalization of the proposed Lipschitz-constrained design to
another one-stage detector, FCOS~\cite{tian2019}. FCOS is an anchor-free detector, using
different mechanisms than SSD such as Feature Pyramid Networks (FPN) and a new
centerness branch. We therefore use this as a test of whether the results observed with
LipSSD can be generalized to other architectures.

\subsection{FCOS}

Instead of relying on predefined anchors, FCOS predicts objects directly from
feature-map locations. In addition to classification and box regression,
\citet{tian2019} introduced a centerness branch, which estimates whether a location lies
near the center of a ground-truth object.

FCOS performs detection over a Feature Pyramid Network (FPN), producing feature maps
($P_3$, $P_4$, $P_5$, $P_6$, $P_7$) with respective strides $(8, 16, 32, 64, 128)$.
Lower-stride levels such as $P_3$ and $P_4$ are used for small objects, while
higher-stride levels such as $P_6$ and $P_7$ target larger objects. Each pyramid level
is processed by two shared towers, one for classification and one for regression. The
classification tower outputs class logits, while the regression tower predicts box
coordinates and centerness scores.

\subsection{LipFCOS architecture}

\begin{figure*}[ht]
    \centering
    \resizebox{\textwidth}{!}{%
        \input{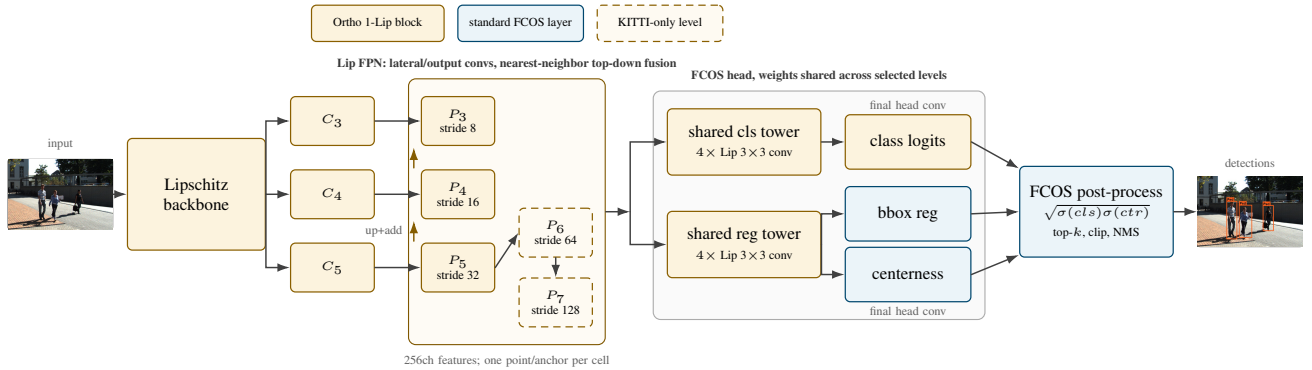}%
    } \caption{\textbf{LipFCOS architecture.} VGG backbone, returned layers $C_3$-$C_5$,
    Lipschitz FPN, and shared FCOS heads. Brown blocks are Lipschitz-constrained; blue
    blocks are standard FCOS layers. The final box-regression and centerness
    convolutions remain unconstrained.}
    \label{fig:lipfcos-architecture}
\end{figure*}

LipFCOS follows the same principle as LipSSD, but applies it to the FCOS backbone, FPN,
and prediction towers. The backbone, FPN, classification tower, regression tower, and
class-logit predictor are replaced with Lipschitz-constrained counterparts. The final
convolutions of the box-regression and centerness branches remain unconstrained,
following the same reasoning as in \cref{sec:method}. A representation of the resulting
architecture with our modifications is shown in \cref{fig:lipfcos-architecture}.

We adapt the original FCOS architecture to KITTI and LARD. For both datasets, we use a
VGG backbone. Since LARD contains only small objects, we remove the $P_6$ and $P_7$
pyramid levels to stabilize training, and only predict on $P_3$, $P_4$, and $P_5$. For
KITTI, the FPN and detection heads are kept unchanged.

\subsection{Experiments}

We train LipFCOS on KITTI and LARD using the original FCOS losses and the same protocol
as in the main experiments. The goal is not to provide an exhaustive FCOS benchmark, but
to test whether the proposed LipFCOS design also improves robustness.

\Cref{tab:map50_results_fcos_safety_critical_attacks} summarizes the results. LipFCOS
comes with a clean-performance cost, but improves robustness under all reported attacks
on both datasets. On KITTI, the clean mAP@50 decreases from $78.7$ to $74.7$, while
attacked performance improves by a large margin across all attack types. On LARD, the
clean drop is larger than with LipSSD, but the robustness gain is also more important.

\begin{table*}[ht]
\centering
\begin{tabular}{llccccccc}
\toprule
Dataset & Model & Clean & $A_{\mathrm{cls}}$ & $A_{\mathrm{reg}}$ & DAG & TOG-F & TOG-V
& TOG-M \\
\midrule
\multirow{2}{*}{KITTI} & FCOS & \textbf{78.7} & 35.1 & 36.9 & 37.3 & 44.1 & 57.5 & 66.7
\\
  & LipFCOS  & 74.7 & \textbf{69.0} & \textbf{70.7} & \textbf{72.8} & \textbf{73.5} &
  \textbf{74.0} & \textbf{74.7} \\
\midrule
\multirow{2}{*}{LARD} & FCOS     & \textbf{96.5} & 16.0 & 2.2 & N/A & 7.5 & 5.2 & N/A \\
  & LipFCOS  & 88.8 & \textbf{67.1} & \textbf{37.4} & N/A & \textbf{87.7} &
  \textbf{82.4} & N/A \\
\bottomrule
\end{tabular}
\caption{mAP@50 for FCOS and LipFCOS on KITTI and LARD under attacks with $\varepsilon =
1.0$.}
\label{tab:map50_results_fcos_safety_critical_attacks}
\end{table*}

\begin{figure*}[ht]
  \centering
  \begin{minipage}[t]{0.48\linewidth}
    \centering
    \resizebox{\linewidth}{!}{
\definecolor{fcosblue}{HTML}{0072B2}
\definecolor{liporange}{HTML}{E69F00}
\begin{tikzpicture}
  \begin{axis}[
      width=\linewidth,
      height=0.66\linewidth,
      xlabel={Attack budget $\varepsilon$},
      ylabel={mAP@50 (\%)},
      xmin=-0.08, xmax=3.08,
      ymin=0, ymax=92,
      xtick={0,0.2,0.5,1,2,3},
      ytick={0,20,40,60,80},
      grid=major,
      major grid style={draw=gray!60, dotted},
      tick align=outside,
      axis x line*=bottom,
      axis y line*=left,
      xtick pos=left,
      ytick pos=left,
      legend pos=north east,
      legend cell align={left},
      legend style={draw=gray!40, font=\small, fill=white,
                    fill opacity=0.9, text opacity=1},
      label style={font=\small},
      tick label style={font=\small},
      every axis plot/.append style={line width=1.1pt, mark size=2.4pt},
  ]
    \addplot[fcosblue, mark=*]
      coordinates {(0,78.70) (0.2,68.56) (0.5,51.53) (1,35.10) (2,20.47) (3,14.14)};
    \addlegendentry{Classic FCOS}
    \addplot[liporange, mark=square*]
      coordinates {(0,74.70) (0.2,73.70) (0.5,71.99) (1,69.00) (2,62.33) (3,54.36)};
    \addlegendentry{LipFCOS}
  \end{axis}
\end{tikzpicture}}
  \end{minipage}\hfill
  \begin{minipage}[t]{0.48\linewidth}
    \centering
    \resizebox{\linewidth}{!}{
\definecolor{fcosblue}{HTML}{0072B2}
\definecolor{liporange}{HTML}{E69F00}
\begin{tikzpicture}
  \begin{axis}[
      width=\linewidth,
      height=0.66\linewidth,
      xlabel={Attack budget $\varepsilon$},
      ylabel={mAP@50 (\%)},
      xmin=-0.08, xmax=1.08,
      ymin=0, ymax=102,
      xtick={0,0.2,0.5,1},
      ytick={0,20,40,60,80,100},
      grid=major,
      major grid style={draw=gray!60, dotted},
      tick align=outside,
      axis x line*=bottom,
      axis y line*=left,
      xtick pos=left,
      ytick pos=left,
      legend pos=north east,
      legend cell align={left},
      legend style={draw=gray!40, font=\small, fill=white,
                    fill opacity=0.9, text opacity=1},
      label style={font=\small},
      tick label style={font=\small},
      every axis plot/.append style={line width=1.1pt, mark size=2.4pt},
  ]
    \addplot[fcosblue, mark=*]
      coordinates {(0,98.90) (0.2,87.38) (0.5,50.65) (1,16.00)};
    \addlegendentry{Classic FCOS}
    \addplot[liporange, mark=square*]
      coordinates {(0,95.55) (0.2,92.51) (0.5,85.39) (1,67.10)};
    \addlegendentry{LipFCOS}
  \end{axis}
\end{tikzpicture}}
  \end{minipage}
  \caption{mAP@50 under $A_{\mathrm{cls}}$ as the attack budget $\varepsilon$ increases
  for KITTI (left) and LARD (right). Note that, for the LARD dataset, the reported
  mAP@50 is evaluated on only 1,000 samples, resulting in a clean mAP@50 that differs
  from the value reported in \cref{tab:map50_results_fcos_safety_critical_attacks}.}
  \label{fig:kitti_fcos_map50_vs_eps}
\end{figure*}
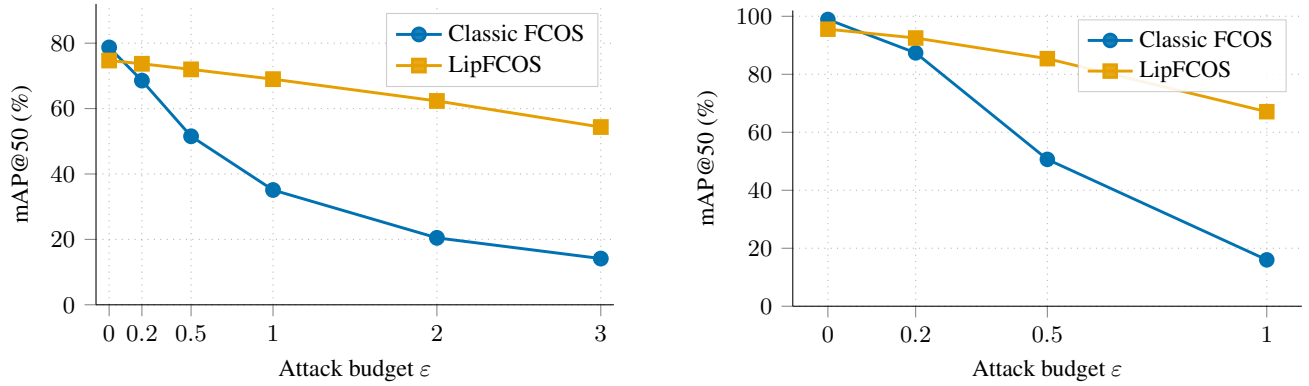

To complement this fixed-budget comparison, \cref{fig:kitti_fcos_map50_vs_eps} reports
mAP@50 on LARD and KITTI as the budget of the $A_{\mathrm{cls}}$ attack increases. This
curve shows the degradation trend that is not visible from a single value at
$\varepsilon = 1.0$. FCOS loses performance faster than the LipFCOS as the budget
increases, following the expected pattern.

\section{LipSSD additional results}

This section complements the main LipSSD experiments with additional quantitative and
qualitative analyses. First, we report degradation curves and larger-budget results to
show how robustness evolves beyond the single-budget comparisons presented in the main
paper. Second, we examine whether the accuracy-robustness trade-off induced by
temperature scaling also appears under a regression-targeted attack. Finally, we
visualize perturbations produced by fabrication attacks, where the objective is to
create false positives rather than suppress existing detections.

\subsection{Performance degradation for varying attacks and budgets}
\label{app:lipssd_quant}

\begin{figure}[ht]
  \centering
  \resizebox{0.52\linewidth}{!}{
\definecolor{ssdblue}{HTML}{0072B2}
\definecolor{liporange}{HTML}{E69F00}
\definecolor{sliporange}{HTML}{D55E00}
\begin{tikzpicture}
  \begin{axis}[
      width=\linewidth,
      height=0.66\linewidth,
      xlabel={Attack budget $\varepsilon$},
      ylabel={mAP@50 (\%)},
      xmin=-0.05, xmax=1.05,
      ymin=0, ymax=102,
      xtick={0,0.2,0.5,1},
      ytick={0,20,40,60,80,100},
      grid=major,
      major grid style={draw=gray!60, dotted},
      tick align=outside,
      axis x line*=bottom,
      axis y line*=left,
      xtick pos=left,
      ytick pos=left,
      legend pos=south west,
      legend cell align={left},
      legend style={draw=gray!40, font=\scriptsize, fill=white,
                    fill opacity=0.9, text opacity=1},
      label style={font=\small},
      tick label style={font=\small},
      every axis plot/.append style={line width=1.1pt, mark size=2.4pt},
  ]
    \addplot[ssdblue, mark=*]
      coordinates {(0,97.30) (0.2,70.67) (0.5,16.88) (1,1.60)};
    \addlegendentry{SSD}
    \addplot[sliporange, mark=triangle*]
      coordinates {(0,97.76) (0.2,90.87) (0.5,52.84) (1,15.40)};
    \addlegendentry{SLipSSD}
    \addplot[liporange, mark=square*]
      coordinates {(0,99.14) (0.2,93.23) (0.5,63.49) (1,20.30)};
    \addlegendentry{LipSSD}
  \end{axis}
\end{tikzpicture}} \caption{mAP@50
  on LARD under $A_{\mathrm{reg}}$ for attack budgets up to $\varepsilon = 1$. Note
  that, the reported mAP@50 is evaluated on only 1,000 samples, resulting in a clean
  mAP@50 that differs from the value reported in
  \cref{tab:map50_results_safety_critical_attacks}}
  \label{fig:lard_map50_vs_eps}
\end{figure}
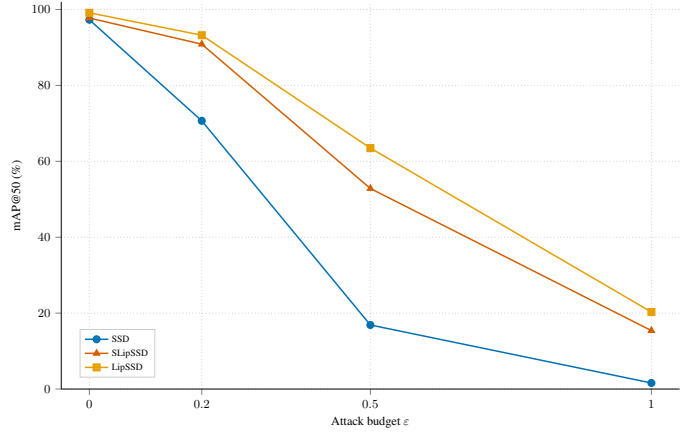

\Cref{fig:lard_map50_vs_eps} reports the $A_{\mathrm{reg}}$ attack on LARD over a budget
$\varepsilon$ range from 0 to 1. This experiment extends results in
\Cref{tab:map50_results_safety_critical_attacks} where only $\varepsilon = 1.0$ is
shown. The curve in \Cref{fig:lard_map50_vs_eps} shows how quickly each model degrades
as the perturbation size increases. The standard SSD loses most of its detection
performance at small budgets. LipSSD and SLipSSD also degrade, but the decrease is more
gradual, and they remain above the vanilla model over the tested range. Thus, the
robustness gain is not restricted to one operating point: it is reflected in the slope
of the degradation curve.

\Cref{tab:map50_results_kitti_eps3_appendix} reports KITTI results for several attacks
under a larger budget $\varepsilon = 3.0$. This setting is useful because it reduces the
ambiguity between clean performance and robustness. At low budgets, SLipSSD remains
competitive on some attacks: even if the drop in performance is higher for SLipSSD, the
mAP@50 under attack is still better because it starts from a higher clean mAP@50. At
$\varepsilon = 3.0$ (see \Cref{tab:map50_results_kitti_eps3_appendix}), this
clean-performance advantage is no longer sufficient, and LipSSD obtains the best mAP@50
under every attack.

\begin{table*}[ht]
\centering
\begin{tabular}{llccccccc}
\toprule
Dataset & Model & Clean & $A_{\mathrm{cls}}$ & $A_{\mathrm{reg}}$ & DAG & TOG-F & TOG-V
& TOG-M \\
\midrule
\multirow{3}{*}{KITTI} & SSD     & 79.8 & 7.6 & 7.0 & 4.8 & 5.7 & 1.6 & 13.4 \\
  & SLipSSD & \textbf{80.0} & 25.5 & 34.4 & 30.3 & 25.0 & 43.2 & 47.8 \\
  & LipSSD  & 76.8 & \textbf{36.4} & \textbf{48.9} & \textbf{55.1} & \textbf{46.3} &
  \textbf{64.9} & \textbf{72.8} \\
\bottomrule
\end{tabular}
\caption{mAP@50 for SSD, SLipSSD, and LipSSD on KITTI under attacks with $\varepsilon =
3.0$.}
\label{tab:map50_results_kitti_eps3_appendix}
\end{table*}

This comparison clarifies the respective roles of the two variants. SLipSSD is the
higher-accuracy variant and provides a favorable compromise when preserving clean
performance is important. LipSSD is the more robust variant, and the difference becomes
clearer as the attack budget increases (see \Cref{fig:kitti_map50_vs_eps}).

\subsection{Robustness to regression-targeted attack}

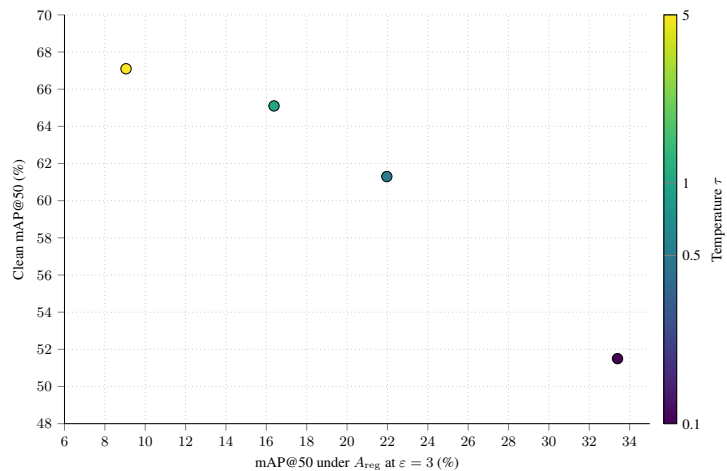
\begin{figure}[ht]
    \centering
    \resizebox{0.55\linewidth}{!}{
\begin{tikzpicture}
  \begin{axis}[
      width=0.82\linewidth,
      height=0.6\linewidth,
      xlabel={mAP@50 under $A_{\mathrm{reg}}$ at $\varepsilon=3$ (\%)},
      ylabel={Clean mAP@50 (\%)},
      xmin=6, xmax=35,
      ymin=48, ymax=70,
      grid=major,
      major grid style={draw=gray!60, dotted},
      tick align=outside,
      axis x line*=bottom,
      axis y line*=left,
      xtick pos=left,
      ytick pos=left,
      label style={font=\small},
      tick label style={font=\small},
      colormap/viridis,
      colorbar,
      point meta min=-1,
      point meta max=0.69897,
      colorbar style={
          ylabel={Temperature $\tau$},
          ytick={-1,-0.30103,0,0.69897},
          yticklabels={0.1,0.5,1,5},
          width=0.28cm,
          ylabel style={font=\small},
          yticklabel style={font=\small},
      },
      scatter/use mapped color={draw=black, fill=mapped color},
  ]
    \addplot[
        scatter, only marks, scatter src=explicit,
        mark=*, mark size=3.2pt,
    ]
      coordinates {
        (9.05,67.1) [0.69897]
        (16.38,65.1) [0]
        (21.97,61.3) [-0.30103]
        (33.40,51.5) [-1]
      };
  \end{axis}
\end{tikzpicture}}
    \caption{Accuracy-robustness trade-off on Pascal VOC for the SLipSSD models trained
    with different temperatures $\tau$. Each point reports clean mAP@50 on full test
    set, vertical axis, against robust mAP@50 under the $A_{\mathrm{reg}}$ attack at
    $\varepsilon = 3$, horizontal axis. The persistence of the trade-off under a
    regression-targeted attack suggests that temperature scaling affects detector-level
    robustness, not only classification robustness.}
    \label{fig:voc_areg_tradeoff}
\end{figure}

Additionally, we investigate whether temperature scaling only improves robustness to
attacks targeting the classification loss (e.g. $A_{\mathrm{cls}}$), or whether it
improves the robustness on other attacks. To this end, we complement the study presented
in~\cref{subsubsec:ar_tradeoff} with the same analysis under the $A_{\mathrm{reg}}$
attack, since the TOG vanishing attack targets the global SSD loss and may focus on the
classification part. \cref{fig:voc_areg_tradeoff} shows the Pareto front for clean
mAP@50 vs. mAP@50 under $A_{\mathrm{reg}}$ attack. Interestingly, the
accuracy-robustness trade-off remains visible under this attack indicating that the
scaling parameter also impacts the detector robustness beyond the classification
objective alone. This observation is consistent with prior work suggesting that
adversarial robustness is closely tied to the features learned by the model, namely
robust and non-robust features~\cite{ilyas2019adversarial}. This is also aligned with
the recent work of~\citet{li2025importance}, showing the backbone's importance in the
adversarial robustness of object detectors. In this view, the effect of temperature
scaling on the $A_\mathcal{\mathrm{reg}}$ attack suggest that it may influence
detector-level robustness through changes in the learned representation by the backbone
rather than only through the classification objective.

\subsection{Visualization of fabrication attack perturbations}

\begin{figure*}[ht]
    \centering
    \input{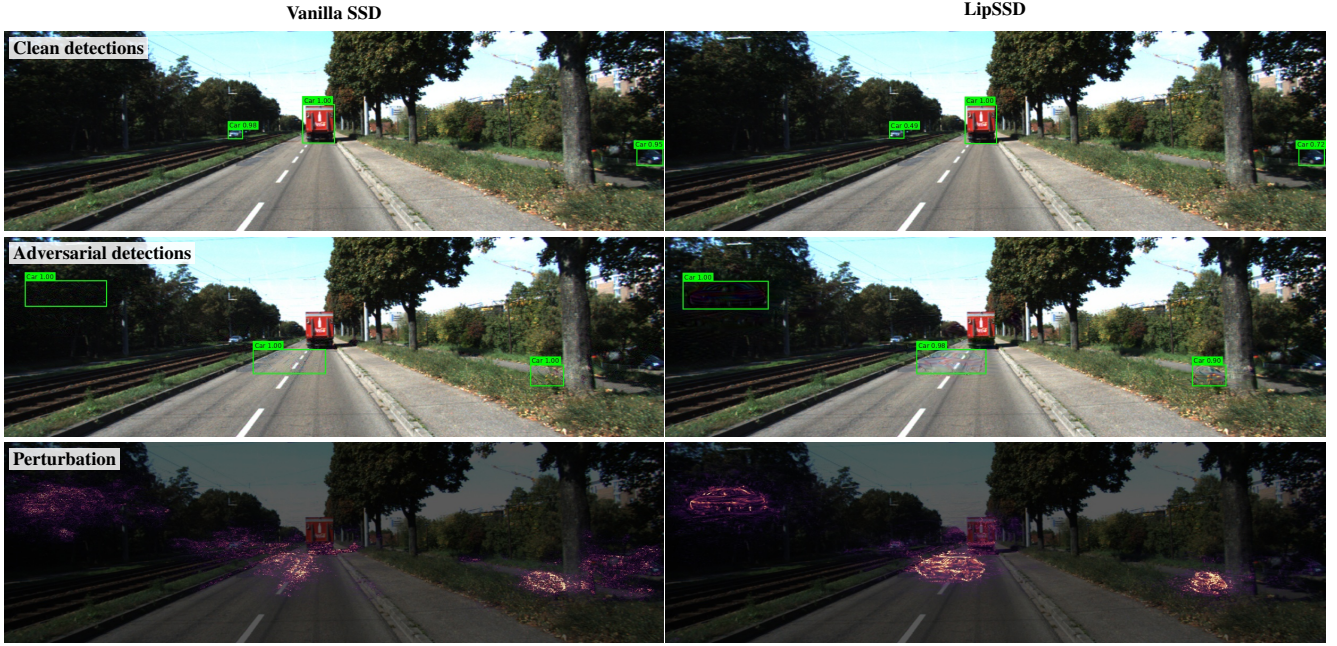}
    \caption{Qualitative comparison under a fabrication objective on KITTI.}
    \label{fig:kitti_tog_fabrication_qualitative}
\end{figure*}

\begin{figure}[ht]
    \centering
   \input{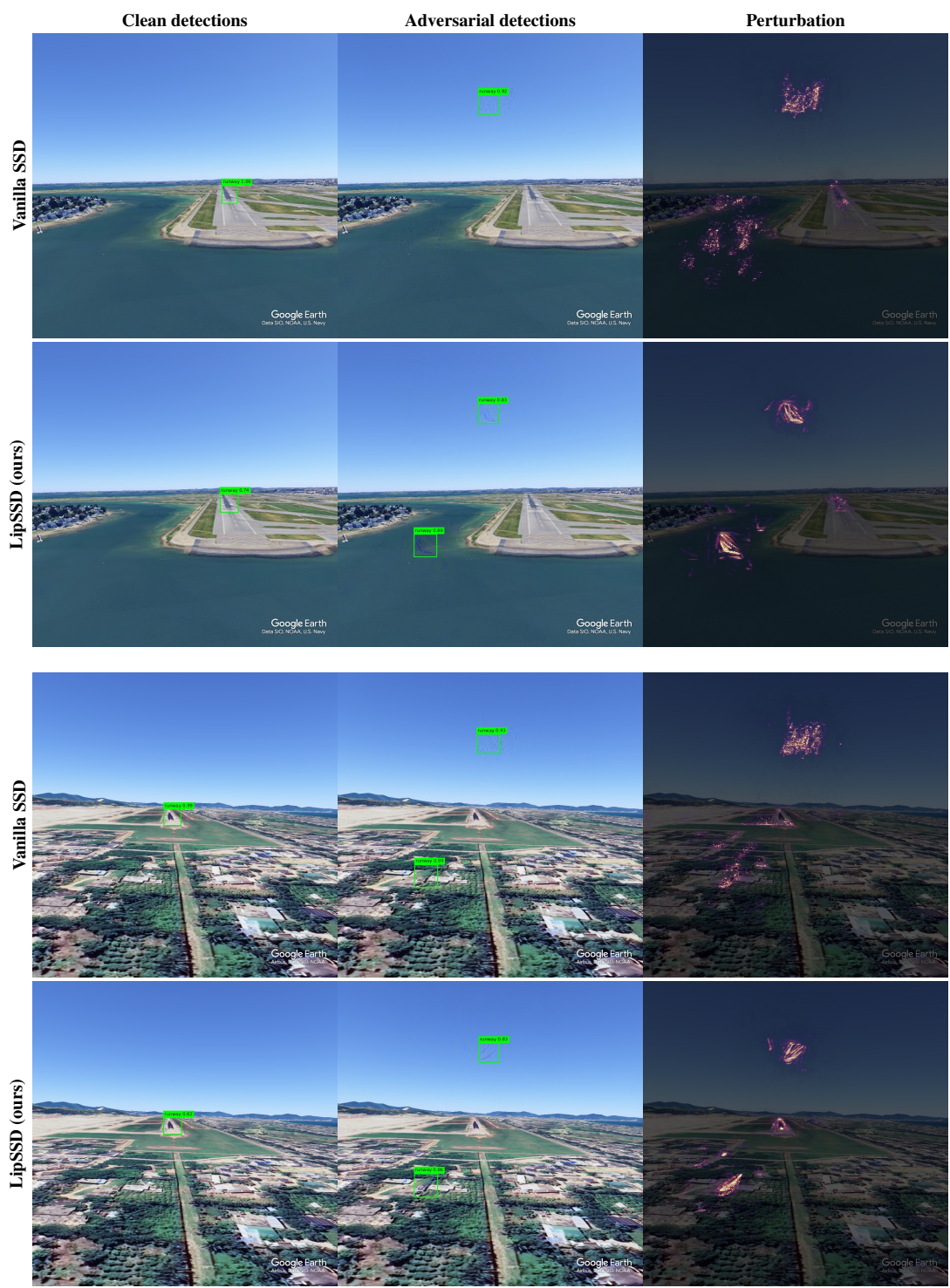}
   \caption{Qualitative comparison under a fabrication objective on LARD.}
    \label{fig:lard_tog_fabrication_qualitative}
\end{figure}

\Cref{fig:kitti_tog_fabrication_qualitative,fig:lard_tog_fabrication_qualitative}
provide qualitative examples under a fabrication objective. Instead of showing whether
detections are preserved under attack, we focus on the perturbation patterns leading to
false positives.

For the vanilla SSD, the fabrication objective can create false detections using mostly
unstructured perturbations. Against LipSSD, the successful perturbations
are visually more organized and tend to align with image structures associated with the
target object. On KITTI, the perturbation is concentrated around car-like structures. On
LARD, the perturbation more often follows runway-like structures. This does not
constitute a formal statement about all possible perturbations, but it illustrates a
qualitative change in the attack solutions found against the Lipschitz-constrained
detector.

\clearpage

\section{Related work on certification of object detectors}

The main paper focuses on empirical robustness under white-box attacks. We discuss here
the related, but distinct, line of work on certified robustness for object detection.
Certified defenses aim to provide formal guarantees, usually called certificates, that a
prediction or performance criterion remains stable within a specified perturbation set.
Such guarantees can be obtained through formal verification, randomized smoothing, or
Lipschitz-based certificates.

Extending certification from classification to detection is difficult because the output
is not a single class label. Detectors produce a variable number of boxes, class scores,
and regression outputs, followed by confidence thresholding and non-maximum suppression.
A certificate must therefore account not only for class stability, but also for
localization, matching, and post-processing effects.

\paragraph{Certified object detection:}
Recent work has started to address certification for object detection. For example,
\citet{cohen2025verifiou} proposed to formally verify the localization robustness of
single-object detectors by bounding the worst-case IoU over a perturbation set. More
recently, \citet{nirala2025} trained and certified a modified YOLOv2-based runway
detector with IBP on 128$\times128$ LARD~\cite{ducoffe2023lard} crops under
$\ell_\infty$ perturbations. These formal verification works remain limited in scope and
are not yet applicable to large detection models and datasets.

In a different line of work, \citet{chiang2020} proposed \textit{Median Smoothing}, a
variant of randomized smoothing for object detection that provides guarantees on both
box coordinates and object classes under an $\ell_2$-norm bounded budget. This approach
is attractive because it directly targets the structured output of object detectors,
rather than reducing detection to a classification problem. However, it is
computationally expensive at inference time: the smoothed detector requires many forward
passes, with the authors using 2000 inferences to perform one smoothed detection. This
makes such methods difficult to deploy in settings where detection latency is critical.

Our work does not provide a certified detector in this sense. Instead, it studies
whether Lipschitz-constrained components can improve empirical robustness while keeping
the standard detection pipeline largely unchanged. This makes the approach closer to a
practical robust-design strategy than to a complete certification method, but it remains
connected to certification through the explicit control of Lipschitz constants.

\end{document}